%% file: main.tex
\documentclass[letterpaper]{article} % DO NOT CHANGE THIS
\usepackage{aaai25}  % DO NOT CHANGE THIS
\usepackage{times}  % DO NOT CHANGE THIS
\usepackage{helvet}  % DO NOT CHANGE THIS
\usepackage{courier}  % DO NOT CHANGE THIS
\usepackage[hyphens]{url}  % DO NOT CHANGE THIS
\usepackage{graphicx} % DO NOT CHANGE THIS
\usepackage{natbib}  % DO NOT CHANGE THIS AND DO NOT ADD ANY OPTIONS TO IT
\usepackage{caption} % DO NOT CHANGE THIS AND DO NOT ADD ANY OPTIONS TO IT
\usepackage{algorithm}
\usepackage{algpseudocode}
\usepackage{booktabs}
\usepackage{multirow}
\usepackage{amssymb}
\usepackage{amsmath}  
\usepackage{amsfonts}
\usepackage[capitalise]{cleveref}
\usepackage{colortbl}
\usepackage{xcolor}
\graphicspath{{figures/}}

\usepackage{enumitem}
\usepackage{amssymb} % For checkboxes

\newlist{checklist}{itemize}{1}
\setlist[checklist,1]{
  label={\(\square\)},
  labelwidth=1em,
  labelsep=1em,
  left=0pt,
  itemsep=0pt
}

\newsavebox\CBox
\def\textBF#1{\sbox\CBox{#1}\resizebox{\wd\CBox}{\ht\CBox}{\textbf{#1}}}

\usepackage{newfloat}
\usepackage{listings}
\DeclareCaptionStyle{ruled}{labelfont=normalfont,labelsep=colon,strut=off} % DO NOT CHANGE THIS
\floatstyle{ruled}
\newfloat{listing}{tb}{lst}{}
\floatname{listing}{Listing}
\title{Towards a Comprehensive, Efficient and Promptable Anatomic Structure Segmentation Model using 3D Whole-body CT Scans}
\author{
    Heng Guo\textsuperscript{\rm 1,2},
    Jianfeng Zhang\textsuperscript{\rm 1,2},
    Jiaxing Huang\textsuperscript{\rm 1},
    Tony C. W. Mok\textsuperscript{\rm 1,2},\\
    Dazhou Guo\textsuperscript{\rm 1},
    Ke Yan\textsuperscript{\rm 1,2},
    Le Lu\textsuperscript{\rm 1},
    Dakai Jin\textsuperscript{\rm 1},
    Minfeng Xu\textsuperscript{\rm 1,2}
}
\affiliations{
    \textsuperscript{\rm 1}DAMO Academy, Alibaba Group\\
    \textsuperscript{\rm 2}Hupan Lab, 310023, Hangzhou, China\\ 
    gh205191@alibaba-inc.com
}

\begin{document}

\maketitle

\input{sec/0_abstract}
\input{sec/1_intro}
\input{sec/2_related_work}
\input{sec/3_method}
\input{sec/4_exp}

\input{sec/5_conclusion}
\bibliography{mybib}

\input{sec/X_suppl}

\end{document}

%% file: sec/0_abstract.tex
\begin{abstract}
Segment anything model (SAM) demonstrates strong generalization ability on natural image segmentation. However, its direct adaptation in medical image segmentation tasks shows significant performance drops. It also requires an excessive number of prompt points to obtain a reasonable accuracy. Although quite a few studies explore adapting SAM into medical image volumes, the efficiency of 2D adaptation methods is unsatisfactory and 3D adaptation methods are only capable of segmenting specific organs/tumors. In this work, we propose a comprehensive and scalable 3D SAM model for whole-body CT segmentation, named CT-SAM3D. Instead of adapting SAM, we propose a 3D promptable segmentation model using a (nearly) fully labeled CT dataset. To train CT-SAM3D effectively, ensuring the model's accurate responses to  higher-dimensional spatial prompts is crucial, and 3D patch-wise training is required due to GPU memory constraints. Therefore, we propose two key technical developments: 1) a progressively and spatially aligned prompt encoding method to effectively encode click prompts in local 3D space; and 2) a cross-patch prompt scheme to capture more 3D spatial context, which is beneficial for reducing the editing workloads when interactively prompting on large organs. CT-SAM3D is trained using a curated dataset of 1204 CT scans containing 107 whole-body anatomies and extensively validated using five datasets, achieving significantly better results against all previous SAM-derived models. Code, data, and our 3D interactive segmentation tool with quasi-real-time responses are available at https://github.com/alibaba-damo-academy/ct-sam3d.

%\keywords{CT-SAM3D \and Comprehensive and Efficient Interactive Segmentation \and 3D Click Based Prompt}

\end{abstract}

%% file: sec/1_intro.tex
\section{Introduction}
\label{sec:intro}

\begin{figure}[t]
\centering
\includegraphics[width=0.9\linewidth]{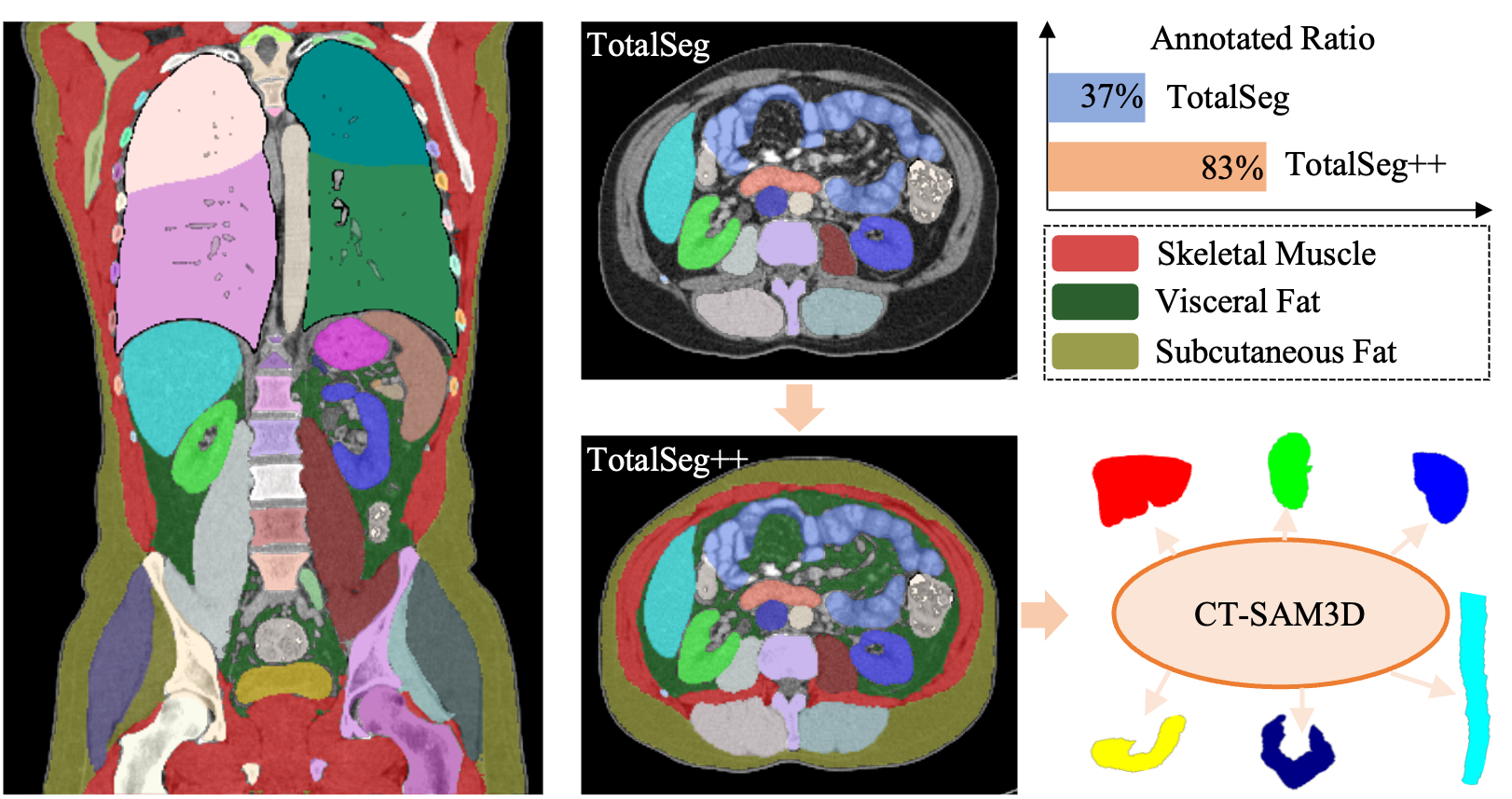}
\caption{Illustration of the enhanced TotalSeg++ dataset and the versatile 3D promptable CT-SAM3D model. TotalSeg++ complements TotalSeg dataset with added skeletal muscle, visceral and subcutaneous fat annotations.}
\label{fig:poc}
\vspace{-1em}
\end{figure}

Image segmentation is a fundamental task in medical image analysis, with ubiquitous clinical applications such as disease quantification \cite{iyer2016quantitative,ferre2019lymphocyte}, computer-aided diagnosis \cite{roth2015improving,chilamkurthy2018deep,mitani2020detection,mckinney2020international}, and radiotherapy planning~\cite{jin2021deeptarget,ye2022comprehensive,jin2022towards}. Despite significant improvements achieved by automatic segmentation methods over the past decade~\cite{wachinger2018keypoint, isensee2021nnu, guo2024med}, it remains challenging in daily clinical use due to large variations in medical images, including different imaging protocols, imaging noises/artifacts, and abnormalities or pathological changes among patients~\cite{hesamian2019deep, albadawy2018deep}. Interactive segmentation or intelligent image editing with human-in-the-loop techniques~\cite{maleike2009interactive,zhao2013overview,sakinis2019interactive,wang2018interactive,wang2019interactive,ji2019scribble,zhang2021interactive,koohbanani2020nuclick} are still needed to further refine the segmentation results. Recently, segment anything model (SAM)~\cite{Kirillov_2023_ICCV} shows great success for general-purpose promptable object segmentation in natural images with strong generalization ability and efficient human interaction. Direct deployment of SAM to medical imaging domain exhibits significant performance drops~\cite{wald2023sam,maciej2023segment,he2023accuracy,huang2023segment,deng2023segment}, but its core design principles of class-agnostic segmentation, prompt encoding, and iterative training scheme can be further exploited and applied to boost the efficiency and accuracy of interactive medical image segmentation.

A few recent studies attempt to fine-tune SAM by incorporating lightweight plug-and-play adapters~\cite{cheng2023sammed2d,chen2023ma,zhang2023customized,yue2023surgicalsam,wu2023medical}. Simple 2D adaptation methods that completely ignore the intrinsic 3D information require numerous clicks when segmenting hundreds of 2D CT slices, rendering them inapplicable in real clinical practice. 
In contrast, other efforts focus on 3D adaptation by integrating a set of 3D adapters into the SAM architecture. 
However, these methods have primarily reported segmentation for a limited number of organs/tumors, and their generalizability to a larger set of 3D anatomical categories has not been validated.

% Note that a more recent work, SAM-Med3D~\cite{wang2023sammed3d}, simply replaces all 2D operations in the SAM network with 3D operations and trains a 3D SAM network from scratch using $21K$ medical images with $131K$ 3D masks. Another recent work, SegVol~\cite{du2023segvol} incorporates a zoom-out-zoom-in mechanism into 3D SAM development based on $6K$ CT scans and $150K$ segmentation masks. As demonstrated later in our experiments, merely extending SAM's network architecture to 3D and scaling up the dataset can lead to significant inefficiency and unsatisfactory results.

In this work, we aim to develop a 3D promptable segmentation model that can interactively segment nearly all anatomic structures within whole-body CT scans with high accuracy and efficiency. To achieve this, we develop a comprehensive, efficient and 3D promptable network, named CT-SAM3D. 
First of all, we identify several key technical challenges in developing the CT-SAM3D model from scratch. 1) SAM's densely annotated dataset (SA-1B) guarantees that each pixel position in its input space has the opportunity to be positively prompted. It is ideal to have an analogous fully labeled whole-body CT dataset, i.e., each voxel in the valid body region has an anatomic label. Otherwise, some anatomical regions would remain as background, thus not being learned or prompted during training, which limits the model's capability in zero-shot or interactive segmentation scenarios. 2) SAM encodes 2D spatial prompts (points/boxes) by using the sum of one-dimensional random Fourier features~\cite{rahimi2007random, tancik2020fourier} and learned attribute embeddings (positive/negative). However, in full 3D space, this prompt encoding method proves less effective than in 2D.
3) Model's complexity and input data scale can increase dramatically in 3D. 

To solve these challenges, we first enrich our whole-body CT scan dataset based on TotalSeg~\cite{wasserthal2023totalsegmentator} by curating the segmentation masks of three important yet under-explored anatomic structures of skeletal muscle, visceral fat, and subcutaneous fat, as illustrated in~\cref{fig:poc}. This results in a more comprehensive whole-body CT dataset, namely TotalSeg++, where overall $\sim$83\% of voxels within the body region are semantically labeled, substantially increased from the previous 37\% in TotalSeg dataset. Then, we propose a progressively and spatially aligned prompt encoding method to ensure the model responds to the 3D spatial prompts accurately. Lastly, 3D patch-wise training is necessary to effectively train a 3D SAM model. Yet, if simply training on 3D local image patches, the inference efficiency would be reduced and drastically more clicks are needed to capture the whole spatial context when segmenting organs larger than the local patch dimensions. Therefore, we propose a cross-patch prompt scheme to alleviate this problem.

We outline our main contributions as follows:

\begin{itemize}
    \item We present a versatile CT-SAM3D model on whole-body CT scans. It is able to segment hundreds of anatomies and achieves new state-of-the-art interactive segmentation results on various datasets.

    \item We propose two technical novelties: 1) a progressively and spatially aligned prompt encoding method to effectively encode click prompts in local 3D space; 2) a cross-patch prompt scheme to make the local click take effect in a broader spatial context.
    
    \item We develop an interactive 3D visualization and segmentation tool with the direct GPU access for model inference. This reports efficient quasi-real-time 3D interactive segmentation performance for the first time.

    \item We enhance the TotalSeg dataset by adding annotations of three important anatomical structures, resulting in a more comprehensively labeled whole-body CT dataset and facilitating the future research in this field.

\end{itemize}

%% file: sec/2_related_work.tex
\section{Related Work}
\label{sec:related}

% \vspace{-1em}
{\bf Segmentation foundation models.} The emergence of foundation models in natural language processing ~\cite{devlin2018bert,brown2020language} has fostered the development of Vision Foundation Models (VFMs)~\cite{caron2021emerging, radford2021learning, oquab2023dinov2,ramesh2022hierarchical}. Based on transformer architectures and training on large datasets, VFMs have the potential to enhance various downstream tasks and demonstrate strong zero-shot capabilities.
SAM~\cite{Kirillov_2023_ICCV} is the first foundation model for generalized image segmentation, validating its zero-shot capability by segmenting objects in the wild. SegGPT~\cite{wang2023seggpt} introduces a general interface compatible with various segmentation tasks. SEEM~\cite{zou2023segment} offers a unified method using varied prompts to segment and identify objects in images all at once. These methods play an important role in inspiring subsequent works.

%Given the importance of the rapidly evolving field in healthcare, several VFMs specialized in the field of medical imaging have been proposed. 
% Yet, only a handful of VFMs focus on the field of medical imaging~\cite{}. 

% In a later work by Yuan et al.~\cite{yuan2021florence}, Florence benefits from universal visual-language representation and further improves its adaptability to more diverse tasks. Trained using large-scale transformer networks (e.g., ViT-L, Swin-L), DINOv2~\cite{} and DALL$\cdot$E 2~\cite{} demonstrate impressive text-to-image generalizability.

% The use of self-supervised learning enables the model to learn from the available data without the need for external labeling or supervision, thus allowing for a more efficient and cost-effective learning process.

% \noindent {\bf Interactive Medical Image Segmentation}
% With limited medical imaging samples, 
% brief review and conclude: No such impressive generality to unseen targets.

%For example, in comparison to natural images, CT scans appear to be less affected by variations in object scales as humans tend to exhibit a degree of similarity in terms of their anatomical structures and organ sizes.
\noindent{\bf SAM adaptation in medical imaging.} Substantial disparities exist between natural images and medical images~\cite{2016Deep}.
The performance of directly applying SAM to medical imaging varies significantly across different objects, anatomies, and modalities~\cite{huang2023segment}. Considerable efforts have been invested to harness the full potential of SAM in medical imaging.
MedSAM~\cite{ma2024segment} curates a medical image cohort of $200K$ masks and adapts SAM to medical image segmentation.
%Low-rank adaptation~\cite{hu2021lora} greatly reduces the number of trainable parameters for fine-tuning tasks, making adapters popular across various fields. 
SAM-Med2D~\cite{cheng2023sammed2d} and SAMed~\cite{zhang2023customized} use 2D adapters for medical images. MA-SAM~\cite{chen2023ma} and 3DSAM-adapter~\cite{gong20233dsam} incorporate a set of 3D adapters into each transformer block of the encoder to extract 3D information in medical scans. 
%MSA~\cite{wu2023medical} introduces space and depth adapters specifically designed to process 3D information.
However, adaptation methods are often restricted to a limited number of organs/tumors.
Alternatively, training 3D models from scratch can directly capture 3D contexts. SAM-Med3D~\cite{wang2023sammed3d} simply reforms the 2D SAM into its 3D counterpart to train a 3D SAM model from scratch using $21K$ medical images and $131K$ masks. SegVol~\cite{du2023segvol} incorporates a zoom-out-zoom-in mechanism into 3D SAM development based on $6K$ CT scans and $150K$ masks.
Despite the significant increase in images and masks, the challenges associated with developing a 3D SAM model remain unsolved.

%In this work, we develop a 3D promptable model from scratch by enhancing TotalSeg with 1204 CT scans, only $\sim$5\% of data used in SAM-Med3D and $\sim$20\% of that used in SegVol, but with a higher annotated ratio ($\sim$83\%) within valid body region.
%As validated in our extensive experiments, CT-SAM3D significantly outperforms previous SAM-derived models by a large margin while requiring fewer clicks as prompts.

%% file: sec/3_method.tex
\section{Methodology}
\label{sec:method}

\begin{figure*}[t]
\centering
\includegraphics[width=0.85\linewidth]{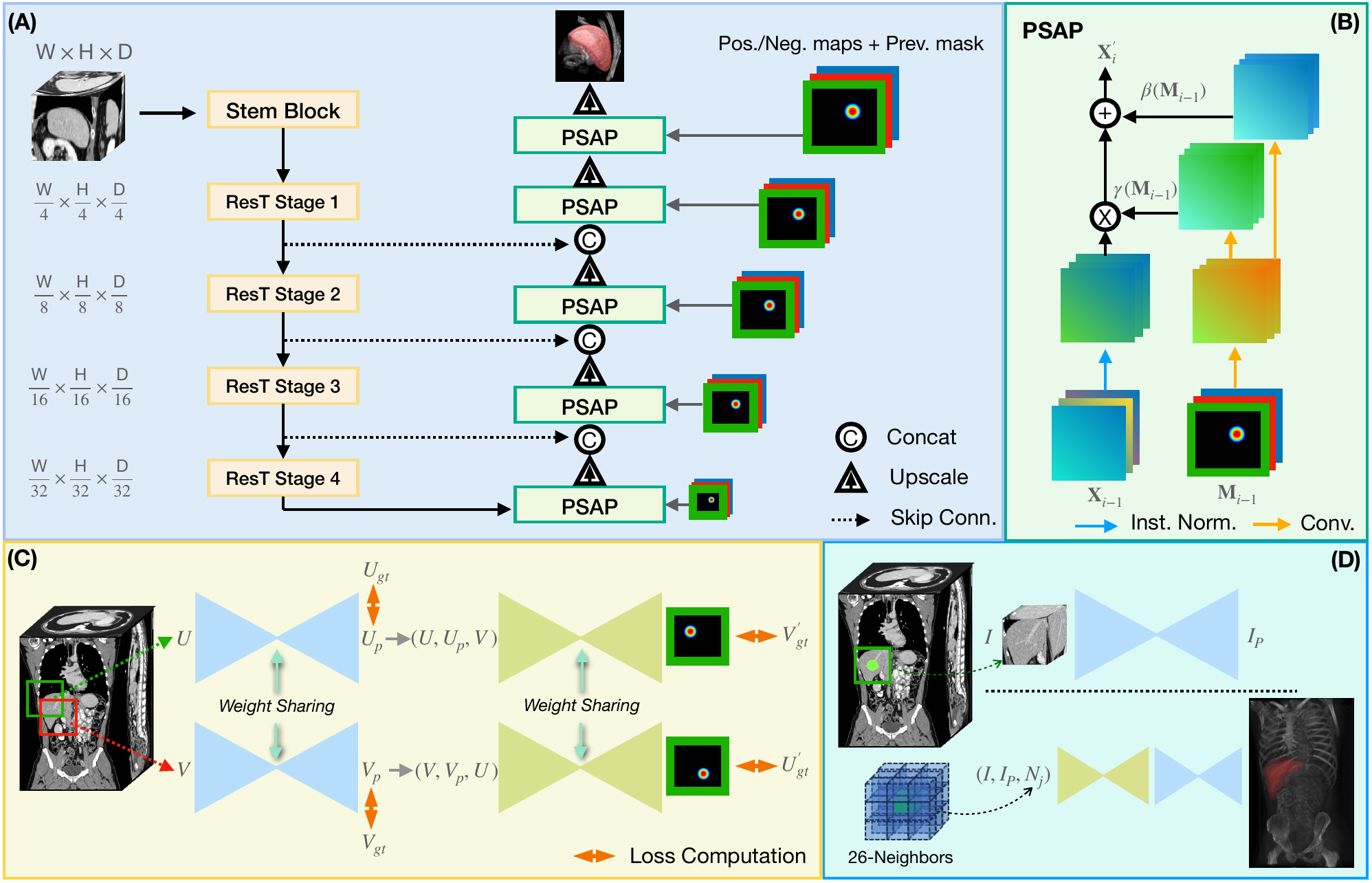}
\caption{\small (A) Framework of CT-SAM3D. (B) Details of progressively and spatially aligned prompt. (C) Cross-patch prompt training scheme. (D) Inference on large organs via cross-patch prompting on $\mathbf{N}_j (j\in [1, 26])$, which are the nearest neighbors around the selected patch.}
\label{fig:arch} \vspace{-1mm}
\end{figure*}

\subsection{CT-SAM3D Architecture}\label{sec:architecture}
Recall that SAM's ViT~\cite{dosovitskiy2020image} backbone employs a convolution kernel $(16 \times 16)$ to patchify the input image $(1024 \times 1024)$, producing a sequence of 4096 tokens.
When dealing with 3D space, we need to avoid the exponential increase in tokens. In terms of feature extraction, it has been extensively shown that hierarchical multi-scale features play a crucial role in semantic segmentation~\cite{ronneberger2015u, cciccek20163d, isensee2021nnu}. A recent SAM-derived study reports that simply dividing the ViT into four stages and establishing connections between the feature maps of each stage and the corresponding decoder layers does not confer any advantages~\cite{chen2023ma}. Taking these into account, we incorporate a hierarchical and memory-efficient Transformer network ResT~\cite{zhang2022rest} and construct skip connections~\cite{ronneberger2015u} to form another U-shape architecture, as illustrated in~\cref{fig:arch}(A). 
The four-stage ResT-based image encoder has a stem block consisting of two consecutive 3D convolution layers as the initial feature extractor.
For a given 3D input patch $I\in \mathbb{R}^{W \times H \times D}$, it will be down-scaled by factors of [4, 8, 16, 32] in the hierarchical encoding path.
The decoding feature integration is achieved by channel-wise concatenation. The 3D transposed convolutions with kernel size of 2 and stride of 2 are used for feature upscaling. The prompt signals are applied to every decoder stage of the network.

\subsection{Progressively and Spatially Aligned Prompt} \label{sec:Aligned}
The vanilla SAM's spatial prompt has showcased a robust capability of encoding 2D positions through the utilization of random Fourier features (RFF)~\cite{rahimi2007random, tancik2020fourier}. Besides learning to predict masks based on these one-dimensional embeddings, the SAM model also needs to learn an additional embedding to distinguish between positive and negative points. Given a prompted point, it will firstly be normalized to a vector $\mathbf{v} \in [-1, 1]^d$ relative to the input image size, where $d = 2$ in SAM's space. Subsequently, a set of sinusoids is generated as:
\begin{equation}
\label{eq:rff}
\begin{aligned}
\textit{RFF}(\mathbf{v})=[&\cos \left(2 \pi \mathbf{b}_1^{\mathrm{T}} \mathbf{v}\right), \sin \left(2 \pi \mathbf{b}_1^{\mathrm{T}} \mathbf{v}\right), \ldots, \\ 
&\cos \left(2 \pi \mathbf{b}_m^{\mathrm{T}} \mathbf{v}\right), \sin \left(2 \pi \mathbf{b}_m^{\mathrm{T}} \mathbf{v}\right)]^{\mathrm{T}},
\end{aligned}
\end{equation}
where $m$ is a configurable feature length, and $\mathbf{b}_j (j\in [1, m])$ is sampled from an isotropic distribution. The final prompt encoding of $\mathbf{v}$ is $\textit{PE}(\mathbf{v})=\textit{RFF}(\mathbf{v}) + \mathbf{e}$, where $\mathbf{e}$ is the learned positive/negative embedding. We experimentally found that simply adapting this technique to 3D is less effective than in 2D, as the spatial alignment between the prompt embedding and the 3D position is more challenging to learn. This may lead to unexpected interactive behavior.

Partially inspired by the spatially-adaptive normalization~\cite{park2019semantic} that effectively preserves the location geometry of the semantic label map, we propose a progressively and spatially aligned prompt (PSAP) for 3D prompt encoding. In contrast to SAM's prompt encoding, our proposed method encodes positive and negative points into two separate click maps, eliminating the need to learn the positive/negative attribute embedding for prompts. This principle is also adopted by some previous 2D interactive segmentation methods~\cite{xu2016deep, sofiiuk2022reviving, chen2022focalclick, liu2023simpleclick}. 
Our PSAP differs in that it omits the need for forwarding through a computationally intensive image encoder, ensuring faster interactive responses. Concretely, 
for a given click point $\mathbf{v}=(x, y, z)$, a Gaussian heatmap is generated around this point. This heatmap is regarded as feature map $\mathbf{P}$ if it is a positive click; otherwise, it is feature map $\mathbf{N}$. Subsequently, the mask prediction $\mathbf{Y}$ from the previous iteration (filled with zeros in the initial iteration), is concatenated with $\mathbf{P}$ and $\mathbf{N}$ to form the composite feature map $\mathbf{M}$ with the order of $[\mathbf{P}, \mathbf{N}, \mathbf{Y}]$. Then, let $\mathbf{X}_{i-1}$ denote the feature map output of the $(i-1)^{th}$ decoder layer, $\mathbf{M}$ will be downsampled to $\mathbf{M}_{i-1}$ that has the same spatial dimensions as $\mathbf{X}_{i-1}$. As illustrated in~\cref{fig:arch}(B), we compute $\gamma(\mathbf{M}_{i-1}) = Conv_{\gamma}(Conv(\mathbf{M}_{i-1}))$ and $\beta(\mathbf{M}_{i-1}) = Conv_{\beta}(Conv(\mathbf{M}_{i-1}))$, then the output of the proposed PSAP is formulated as follows: 
\begin{equation}
\label{psap}
\begin{aligned}
\mathbf{X}_{i}^{'} = \gamma(\mathbf{M}_{i-1})\text{IN}(\mathbf{X}_{i-1}) + \beta(\mathbf{M}_{i-1}),
\end{aligned}
\end{equation}
where $\gamma(\mathbf{M}_{i-1})$ and $\beta(\mathbf{M}_{i-1})$ are the learned modulation parameters of the instance normalization (IN) layer~\cite{ulyanov2016instance}, and they both have the same spatial dimensions as $\mathbf{X}_{i-1}$. Finally, $\mathbf{X}_{i}^{'}$ will be upscaled and concatenated with the corresponding encoder features to form $\mathbf{X}_{i}$. This technique enables the model to encode 3D clicks along the decoding path progressively and spatially aligned, resulting in improved efficiency and accuracy for 3D interactive segmentation.

\subsection{Cross-Patch Prompt for Large Anatomy} \label{sec:Cross}
Due to limitations in GPU memory and computation efficiency, it is not feasible to take as input an entire 3D volume (e.g., $512\times512\times192$ for a typical Thoracic CT scan) for network training. However, cropping 3D medical images into smaller sub-volumes can result in truncation of larger or tubular-shaped anatomies, e.g., liver or aorta. 
Training on only 3D local image patches reduces inference efficiency and requires more manual clicks. To alleviate this problem, we introduce an innovative cross-patch prompt (CPP) approach to capture a broader spatial context. Specifically, beyond the promptable segmentation model (denoted as $\mathcal{S}$ and condensed as the light blue modules in~\cref{fig:arch}(C)), we add a light-weight encoder-decoder sub-network (denoted as $\mathcal{P}$ and condensed as the light green modules in~\cref{fig:arch}(C)), which is dedicated to predicting prompts for neighboring patches given a clicked patch and its mask prediction. 

To achieve this, we sample two patches with overlapping regions in each anatomy during training iterations. As illustrated in ~\cref{alg:cpp}, the training pipeline is executed in $N$ iterations to simulate a real-world interactive scenario. Let $U$ and $V$ denote two patches sharing overlapping regions, $U_C$ and $V_C$ denote the sampling clicks from the simulated user. The outputs of the segmentation network for the two input patches are $U_p$ and $V_p$, and the associated segmentation loss for two local patches $\mathcal{L}_{local}$ is calculated (line 3), where $\mathcal{L}_{seg}(\cdot)$ denotes the combination of Focal loss~\cite{lin2017focal} and Dice loss~\cite{milletari2016v}. Then, the CPP prediction module takes as inputs the crossed and concatenated tuples $(U, U_p, V)$ and $(V, V_p, U)$ to form a mutual anatomical region identification task (line 4). In this task, we aim to predict a heatmap that can be used as click prompts as aforementioned. The ground truth heatmaps $V_{gt}^{'}$ and $U_{gt}^{'}$ are constructed based on the centroids of the foregrounds in $V_{gt}$ and $U_{gt}$, respectively. 
After obtaining the CPP predictions $U_C^{'}$ and $V_C^{'}$, we compute $\mathcal{L}_{cpp}(\cdot)$ using a mean squared error loss, then the cross-patch loss $\mathcal{L}_{cross}$ is calculated (line 5). The CT-SAM3D is then updated (line 7), considering both the local segmentation loss and cross-patch prompt prediction loss. We omit the mask input and its updating in the pseudo-code for simplicity. This training process is also illustrated in ~\cref{fig:arch}(C).

When segmenting a large anatomy such as the liver or aorta, a straightforward approach involves starting with an initial click to obtain the corresponding mask of the local patch. Then, the number of clicks can be gradually increased in the uncovered areas until the complete result is achieved. Yet, this is not user-friendly. As shown in~\cref{fig:arch}(D), CPP can reduce workloads by utilizing the 26 nearest spatial neighboring patches surrounding the selected patch, thereby attaining precise segmentation results with fewer clicks.

%  we can utilize the initial local patch $I$ and mask prediction $P$ as a support, then CPP out the results of its neighbors. This will significantly reduce the number of clicks. 
% Performing CPP on a larger scale is often unnecessary and can introduce additional computational burden, slowing down the response of the system.

\begin{algorithm}[t]
\small
  \caption{Training Algorithm of CT-SAM3D}  
  \label{alg:cpp}
  \begin{algorithmic}[1]
    \Require  
      \Statex $U, V$: Patch samples with overlapping;
      \Statex $U_{gt}, V_{gt}$: Ground truth masks;
      \Statex $U_{C}, V_{C}$: Sampling clicks;
      \Statex $\epsilon$: Learning rate;
      \Statex $N$: Number of iterations per sample;  
      \Statex $\mathcal{S}, \mathcal{P}$: Segmentation and CPP prediction modules.
    \Ensure  
      Optimal network parameters $\theta$  
    \While{$N > 0$ }
     \State $U_p = \mathcal{S}(U, U_C)$, $V_p = \mathcal{S}(V, V_C)$
     \State $\mathcal{L}_{local} = \mathcal{L}_{seg}(U_p, U_{gt})+\mathcal{L}_{seg}(V_p, V_{gt})$
     \State $V_C^{'} = \mathcal{P}(U, U_p, V), U_C^{'} = \mathcal{P}(V, V_p, U)$
     \State $\mathcal{L}_{cross} = \mathcal{L}_{cpp}(U_C^{'}, U_{gt}^{'})+\mathcal{L}_{cpp}(V_C^{'}, V_{gt}^{'})$
     \State $\mathcal{L} = \mathcal{L}_{local} + \mathcal{L}_{cross}$
     \State $\theta \leftarrow {\theta -\epsilon\nabla_{\theta}{\mathcal{L}}}$
     \State Update $U_C, V_C$ according to the error regions
     \State $N \leftarrow N - 1$
    \EndWhile
  \end{algorithmic}
\end{algorithm}

%% file: sec/4_exp.tex
\section{Experiments}
\label{sec:exp}
We first introduce the datasets used in our experiments.
% We train CT-SAM3D using the training set of the curated TotalSeg++ dataset. Then, the testing set of TotalSeg++ is used as the internal testing set and four public datasets (FLARE22, BTCV, MSD-Pancreas, and MSD-Colon) are treated as external testing sets. 
%SAM and several SAM-derived methods in medical imaging are extensively compared. 
%\subsection{Datasets}

\noindent {\bf TotalSeg} dataset consists of 1204 CT scans with 104 anatomical structures annotated~\cite{wasserthal2023totalsegmentator}. Using an in-house developed muscle/fat segmentation model with manual examination and curation, we enhance the TotalSeg dataset by introducing annotations of three anatomies, i.e., skeletal muscle, visceral fat, and subcutaneous fat. This results in a comprehensive whole-body CT dataset where overall $\sim$83\% of voxels within the body possesses a semantic label. We refer to this enhanced dataset as {\bf TotalSeg++}. Details of the data curation process are in our supplementary material. We follow the original data split, using 1139 CT scans for training and 65 for internal testing.

 \noindent {\bf FLARE22} is proposed in an abdominal organ segmentation challenge held at MICCAI 2022~\cite{ma2023unleashing}. The 13 labeled organs include the liver, spleen, pancreas, etc. The offline test set of 20 CT cases is used for external testing. 

 \noindent {\bf BTCV} is also an abdominal challenge dataset~\cite{landman2015miccai} that includes 30 CT scans with annotations for 13 organs, differing slightly from FLARE22 as it lacks duodenum but includes portal vein and splenic vein annotations. The total 30 CT scans are also used for external testing.

\noindent {\bf MSD-Pancreas} and {\bf MSD-Colon} are two tumor segmentation datasets from Medical Segmentation Decathlon challenge~\cite{antonelli2022medical}. They contain 281 and 126 abdominal CT scans respectively. They are used to validate the model's zero-shot segmentation capability.

% \vspace{2em}
\subsection{Evaluation Protocol}
To maximize the reproducibility, we follow the practice in~\cite{wang2023sammed3d, Kirillov_2023_ICCV, sofiiuk2022reviving} to simulate the real-world interactive scenario. Specifically, the first simulated click point is randomly sampled from the foreground region, and the subsequent point is sampled iteratively using the farthest point from the boundary of error regions. 
%The subsequent point is positive if false negative is bigger than false positive otherwise negative. 
We measure 3D organ-wise Dice Similarity Coefficient (DSC) and Normalized Surface Distance (NSD)~\cite{maier2022metrics} after $N$ point prompts, where $N\in\{1, 3, 5, 7, 9\}$. Hence, predictions from 2D SAM-derived methods need to be heavily merged before calculating the metrics. Note that we take into account the number of prompt points in the corresponding model space for evaluation. In other words, for a 2D SAM-derived model, $N$ indicates the number of prompts used in each 2D CT slice (not multiplied by the number of slices), whereas for a 3D SAM-derived model, $N$ represents the number of prompts in a whole 3D CT scan. 
%This evaluation protocol actually favors the 2D models, since there are a total of $N \times$ organ-slices prompts used to segment an organ. In comparison, only $N$ prompts are utilized to segment an organ by the 3D models.  
Even under this biased evaluation protocol, our experiments demonstrate that CT-SAM3D significantly outperforms all other 2D SAM-derived methods by a large margin of at least $10\%$ DSC, while actually requiring fewer clicks.
%although using a much smaller number of prompts.
%Taking into account the total number of points would unfairly disadvantage 2D methods. Because it does not truly reflect the interactive capabilities of the model, as both 2D and 3D models showcase their interactive abilities by updating predictions based on prompts within their respective input spaces. 

\noindent \textbf{Comparing methods.} SAM~\cite{Kirillov_2023_ICCV} and recent SAM-inspired medical image segmentation models are primarily and extensively compared, including MedSAM~\cite{ma2024segment}, SAMed~\cite{zhang2023customized}, MA-SAM~\cite{chen2023ma}, SAM-Med2D~\cite{cheng2023sammed2d}, SAM-Med3D~\cite{wang2023sammed3d} and SegVol~\cite{du2023segvol}. 
%Except that SAM-Med3D and SegVol are nearly trained from scratch, all other models are fine-tuned or adapted to medical images. 
%All these methods provide the pre-trained networks so that we directly apply their models to our testing datasets to evaluate and report the corresponding quantitative segmentation performances. 
Note that SAMed and MA-SAM disable the prompt encoding module, so only a fixed number of organs that have appeared in their training datasets can be segmented.

% \vspace{-1.5em}
\noindent {\bf Implementation details.} Our implementation is built upon PyTorch \cite{paszke2019pytorch}. CT-SAM3D is trained using AdamW optimizer \cite{loshchilov2017decoupled} with an initial learning rate of $1\mathrm{e}{-4}$. The total training process consists of 1000 epochs, with the first 100 epochs to linearly warm up. The learning rate is then reduced by a factor of 10 at the 800th epoch. Our model is trained on 8 A100 GPUs, with a batch size of 4 per GPU and a sampling number of 8 per volume. The number of iterations per batch is set to be 5. The first iteration only uses points as prompts, and the subsequent iterations use both updated points and previous masks as prompts simultaneously. For $\mathcal{L}_{seg}(\cdot)$ in~\cref{alg:cpp}, we use a combination of Focal loss~\cite{lin2017focal} and Dice loss~\cite{milletari2016v} with coefficients of 0.2 and 0.8, respectively.
For data preprocessing, we process all data to have an isotropic spacing of 1.5 mm and apply normalization to scale it to the range of [0, 1]. The patch size is set as $(64\times64\times64)$. Random cropping is employed to obtain training samples. 
%No additional data augmentation is used either in training or inference stages. 
Notably, CT-SAM3D is trained from scratch. To facilitate the developing and validation of CT-SAM3D, we have developed an interactive 3D visualization and segmentation tool with quasi-real-time responses, details of which are provided in the supplementary.

\begin{figure}[t]
\centering
\includegraphics[width=1.0\linewidth]{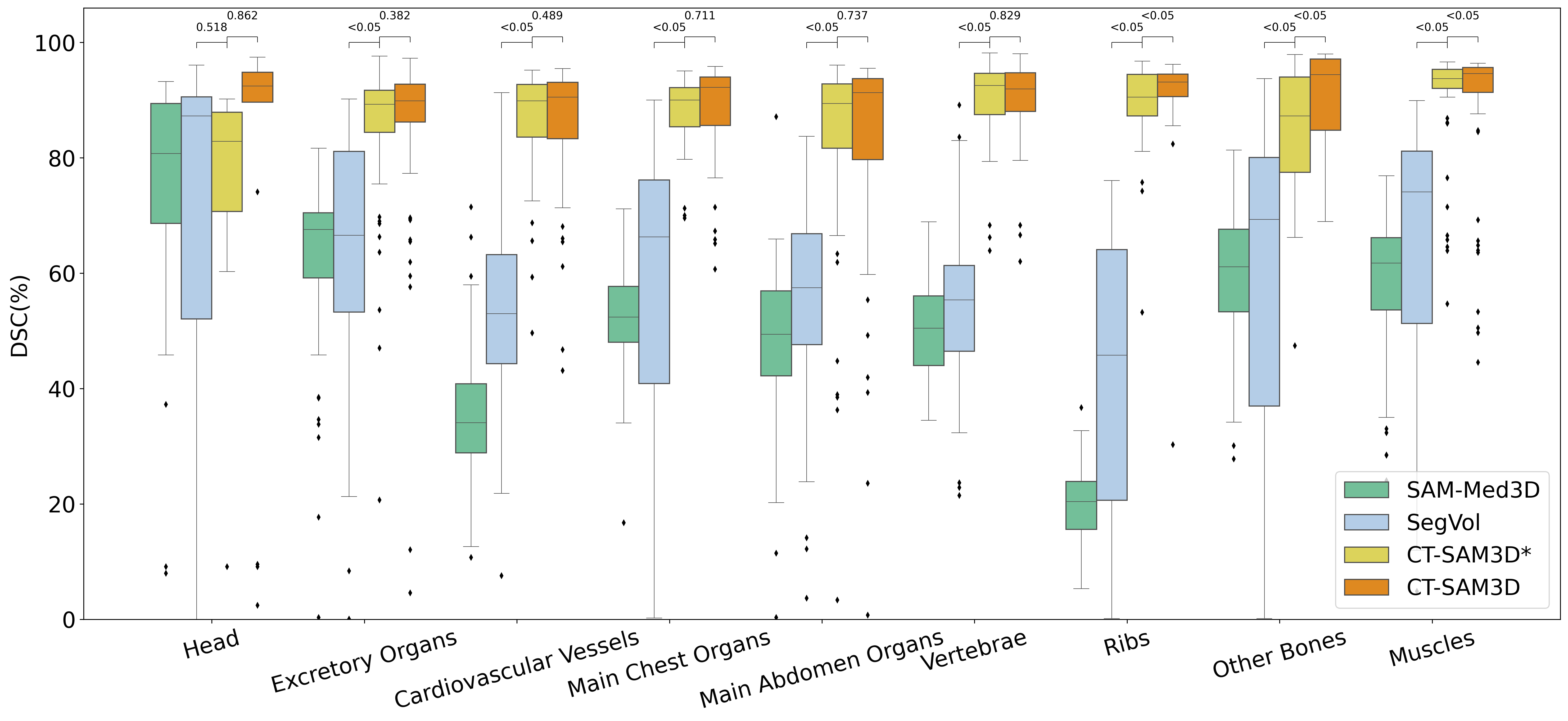}
\vspace{-2em}
\caption{\small Grouped boxplot of different methods. ``CT-SAM3D*'' (\textcolor[rgb]{0.965, 0.882, 0.273}{lemon} color) denotes degraded results when trained on TotalSeg. The $p$-values are presented above the boxes.}
\label{fig:totalseg}
\end{figure}

\begin{table*}[t]
	\centering
	%\addtolength{\leftskip} {-1cm}
	\resizebox{\textwidth}{!}{
	\begin{tabular}{lllllllllllllll}
		\toprule[1pt]
		Methods & Liver & Kidney\_R & Spleen & Pancreas & Aorta & IVC & RAG & LAG & Gallbladder & Esophagus & Stomach & Duodenum & Kidney\_L & mDSC$\uparrow$\\
		\midrule
		SAM & 86.0\scriptsize{$\pm$5.5} & 87.6\scriptsize{$\pm$8.7} & 84.5\scriptsize{$\pm$8.9} & 53.4\scriptsize{$\pm$10.6} & 77.5\scriptsize{$\pm$16.1} & 44.5\scriptsize{$\pm$16.3} & 19.4\scriptsize{$\pm$14.0} & 33.9\scriptsize{$\pm$15.0} & 52.4\scriptsize{$\pm$16.4} & 35.2\scriptsize{$\pm$6.8} & 68.0\scriptsize{$\pm$11.2} & 44.4\scriptsize{$\pm$12.4} & 82.6\scriptsize{$\pm$11.3} & 59.2\\
		  MedSAM & 93.0\scriptsize{$\pm$3.1} & 90.0\scriptsize{$\pm$5.3} & 89.1\scriptsize{$\pm$11.0} & 73.5\scriptsize{$\pm$9.6} & 82.5\scriptsize{$\pm$19.7} & 76.5\scriptsize{$\pm$19.4} & 36.0\scriptsize{$\pm$23.6} & 48.7\scriptsize{$\pm$22.6} & 56.4\scriptsize{$\pm$27.1} & 64.7\scriptsize{$\pm$19.9} & 84.0\scriptsize{$\pm$13.0} & 53.9\scriptsize{$\pm$11.7} & 89.7\scriptsize{$\pm$7.9} & 72.2\\
            SAMed & 92.4\scriptsize{$\pm$5.1} & 74.6\scriptsize{$\pm$28.9} & 90.6\scriptsize{$\pm$6.3} & 65.0\scriptsize{$\pm$19.6} & 83.4\scriptsize{$\pm$11.7} & -- & -- & -- & 71.3\scriptsize{$\pm$25.4} & -- & 76.9\scriptsize{$\pm$21.5} & -- & 77.2\scriptsize{$\pm$27.4} & 78.9*\\        
            MA-SAM & 92.8\scriptsize{$\pm$5.4} & 80.0\scriptsize{$\pm$20.1} & 87.6\scriptsize{$\pm$13.4} & 74.1\scriptsize{$\pm$14.1} & 86.4\scriptsize{$\pm$10.2} & 80.1\scriptsize{$\pm$16.7} & 45.0\scriptsize{$\pm$16.3} &  46.9\scriptsize{$\pm$18.2} & 77.1\scriptsize{$\pm$13.2} & 70.5\scriptsize{$\pm$17.8} & 75.9\scriptsize{$\pm$22.0} & -- & 77.3\scriptsize{$\pm$25.4} & 74.5*\\
            SAM-Med2D & 91.4\scriptsize{$\pm$5.8} & 83.7\scriptsize{$\pm$17.3} & 83.9\scriptsize{$\pm$15.2} & 58.8\scriptsize{$\pm$18.7} & 60.6\scriptsize{$\pm$22.5} & 18.6\scriptsize{$\pm$10.4} & 10.6\scriptsize{$\pm$9.7} & 27.1\scriptsize{$\pm$12.4} & 32.9\scriptsize{$\pm$21.6} & 28.1\scriptsize{$\pm$13.3} & 72.9\scriptsize{$\pm$16.6} & 45.4\scriptsize{$\pm$19.8} & 86.0\scriptsize{$\pm$16.8} & 53.8\\
		
            \midrule
		SAM-Med3D & 85.4\scriptsize{$\pm$13.2} & 84.2\scriptsize{$\pm$9.5} & 84.7\scriptsize{$\pm$11.8} & 46.9\scriptsize{$\pm$14.3} & 60.4\scriptsize{$\pm$10.7} & 44.5\scriptsize{$\pm$13.4} & 32.6\scriptsize{$\pm$20.9} & 35.3\scriptsize{$\pm$18.3} & 56.0\scriptsize{$\pm$19.4} & 32.6\scriptsize{$\pm$16.4} & 46.9\scriptsize{$\pm$19.8} & 27.4\scriptsize{$\pm$13.6} & 84.9\scriptsize{$\pm$6.9} & 55.5\\
            SegVol & 83.9\scriptsize{$\pm$25.3} & 71.7\scriptsize{$\pm$30.6} & 75.9\scriptsize{$\pm$28.8} & 69.4\scriptsize{$\pm$16.1} & 83.1\scriptsize{$\pm$12.1} & 80.3\scriptsize{$\pm$13.9} & 42.1\scriptsize{$\pm$13.3} & 49.7\scriptsize{$\pm$22.6} & 55.6\scriptsize{$\pm$31.1} & 69.6\scriptsize{$\pm$8.4} & 81.1\scriptsize{$\pm$20.7} & 55.6\scriptsize{$\pm$19.8} & 75.1\scriptsize{$\pm$22.6} & 68.7\\
        \rowcolor{gray!20}
		 \textbf{CT-SAM3D} & \textBF{95.6\scriptsize{$\pm$2.0}} & \textBF{95.0\scriptsize{$\pm$1.8}} & \textBF{96.1\scriptsize{$\pm$4.4}} & \textBF{83.6\scriptsize{$\pm$12.0}} & \textBF{94.5\scriptsize{$\pm$2.8}} & \textBF{91.8\scriptsize{$\pm$4.7}} & \textBF{78.4\scriptsize{$\pm$18.0}} & \textBF{82.5\scriptsize{$\pm$4.0}} & \textBF{88.4\scriptsize{$\pm$8.1}} & \textBF{82.9\scriptsize{$\pm$18.1}} & \textBF{92.3\scriptsize{$\pm$4.4}} & \textBF{73.2\scriptsize{$\pm$16.8}} & \textBF{94.8\scriptsize{$\pm$1.4}} & \textBF{88.4}\\
		\bottomrule[1pt]
            \toprule[1pt]

            Methods & Liver & Kidney\_R & Spleen & Pancreas & Aorta & IVC & RAG & LAG & Gallbladder & Esophagus & Stomach & Duodenum & Kidney\_L & mNSD$\uparrow$\\
		\midrule
    	SAM & 61.4\scriptsize{$\pm$9.6} & 73.2\scriptsize{$\pm$18.2} & 67.8\scriptsize{$\pm$15.4} & 57.0\scriptsize{$\pm$11.1} & 77.8\scriptsize{$\pm$20.0} & 33.6\scriptsize{$\pm$13.7} & 40.0\scriptsize{$\pm$22.3} & 53.8\scriptsize{$\pm$15.5} & 46.2\scriptsize{$\pm$20.4} & 42.3\scriptsize{$\pm$9.6} & 50.4\scriptsize{$\pm$13.6} & 46.8\scriptsize{$\pm$13.0} & 62.8\scriptsize{$\pm$21.4} & 54.9\\
     
		  MedSAM & 74.7\scriptsize{$\pm$10.4} & 78.6\scriptsize{$\pm$17.3} & 74.0\scriptsize{$\pm$19.8} & 72.1\scriptsize{$\pm$14.9} & 79.6\scriptsize{$\pm$23.5} & 71.8\scriptsize{$\pm$22.6} & 53.2\scriptsize{$\pm$28.1} & 57.8\scriptsize{$\pm$24.0} & 46.5\scriptsize{$\pm$32.2} & 67.5\scriptsize{$\pm$21.3} & 72.8\scriptsize{$\pm$18.3} & 53.4\scriptsize{$\pm$10.5} & 84.4\scriptsize{$\pm$15.9} & 68.2\\
            SAMed & 86.1\scriptsize{$\pm$9.6} & 74.6\scriptsize{$\pm$27.0} & 89.3\scriptsize{$\pm$11.6} & 77.4\scriptsize{$\pm$17.2} & 87.5\scriptsize{$\pm$11.4} & -- & -- & -- & 81.5\scriptsize{$\pm$22.5} & -- & 72.9\scriptsize{$\pm$18.4} & -- & 75.4\scriptsize{$\pm$25.8} & 80.6*\\      
            MA-SAM & 87.6\scriptsize{$\pm$10.7} & 80.3\scriptsize{$\pm$18.1} & 87.9\scriptsize{$\pm$14.8} & 85.9\scriptsize{$\pm$10.8} & 89.8\scriptsize{$\pm$10.2} & 87.7\scriptsize{$\pm$13.6} & 59.6\scriptsize{$\pm$13.6} & 62.0\scriptsize{$\pm$18.9} & 88.1\scriptsize{$\pm$13.1} & 85.7\scriptsize{$\pm$17.5} & 71.4\scriptsize{$\pm$19.7} & -- & 79.6\scriptsize{$\pm$24.0} & 80.5*\\
            SAM-Med2D & 79.5\scriptsize{$\pm$18.3} & 72.9\scriptsize{$\pm$27.2} & 72.0\scriptsize{$\pm$27.2} & 59.1\scriptsize{$\pm$22.8} & 47.6\scriptsize{$\pm$24.7} & 15.0\scriptsize{$\pm$6.7} & 16.1\scriptsize{$\pm$11.6} & 41.5\scriptsize{$\pm$16.5} & 24.6\scriptsize{$\pm$20.2} & 29.4\scriptsize{$\pm$15.7} & 58.8\scriptsize{$\pm$21.5} & 43.2\scriptsize{$\pm$18.8} & 81.8\scriptsize{$\pm$25.1} & 49.3\\
  
            \midrule
		SAM-Med3D & 76.3\scriptsize{$\pm$19.2} & 82.8\scriptsize{$\pm$11.6} & 84.5\scriptsize{$\pm$14.3} & 59.4\scriptsize{$\pm$14.5} & 70.3\scriptsize{$\pm$8.7} & 62.2\scriptsize{$\pm$11.0} & 66.6\scriptsize{$\pm$28.6} & 70.0\scriptsize{$\pm$23.1} & 76.3\scriptsize{$\pm$12.8} & 58.2\scriptsize{$\pm$17.9} & 48.8\scriptsize{$\pm$19.2} & 40.7\scriptsize{$\pm$15.2} & 83.5\scriptsize{$\pm$10.0} & 67.7\\
            SegVol & 79.7\scriptsize{$\pm$25.0} & 74.5\scriptsize{$\pm$29.4} & 77.1\scriptsize{$\pm$26.6} & 83.3\scriptsize{$\pm$13.6} & 91.7\scriptsize{$\pm$9.1} & 92.2\scriptsize{$\pm$10.4} & 75.6\scriptsize{$\pm$12.8} & 76.6\scriptsize{$\pm$23.2} & 68.9\scriptsize{$\pm$33.4} & 91.1\scriptsize{$\pm$7.1} & 84.3\scriptsize{$\pm$21.6} & 69.6\scriptsize{$\pm$19.7} & 78.4\scriptsize{$\pm$23.8} & 80.2\\
        \rowcolor{gray!20}
		 \textbf{CT-SAM3D} & \textBF{94.7\scriptsize{$\pm$4.9}} & \textBF{98.0\scriptsize{$\pm$2.6}} & \textBF{98.1\scriptsize{$\pm$7.5}} & \textBF{92.5\scriptsize{$\pm$15.4}} & \textBF{98.4\scriptsize{$\pm$3.4}} & \textBF{97.4\scriptsize{$\pm$4.9}} & \textBF{95.2\scriptsize{$\pm$19.3}} & \textBF{99.7\scriptsize{$\pm$0.3}} & \textBF{97.6\scriptsize{$\pm$6.1}} & \textBF{94.3\scriptsize{$\pm$18.5}} & \textBF{96.1\scriptsize{$\pm$6.3}} & \textBF{85.3\scriptsize{$\pm$16.9}} & \textBF{98.4\scriptsize{$\pm$2.3}} & \textBF{95.8}\\
		\bottomrule[1pt]
  
	\end{tabular}}
 \vspace{-0.5em}
 \caption{\upshape \small Organ-specific DSC (\%) and NSD (\%) evaluation on FLARE22. The results were obtained after 5 clicks. Abbreviations: ``IVC''-Inferior Vena Cava, ``RAG''-Right Adrenal Gland, ``LAG''-Left Adrenal Gland, ``mDSC''-mean DSC, ``mNSD''-mean NSD. The asterisk symbol (*) signifies that the result gathers exclusively valid values.}
	\label{tbl:quant_flare}
 \vspace{-0.5em}
\end{table*}

\begin{figure}[t]
\centering
\includegraphics[width=1.0\linewidth]{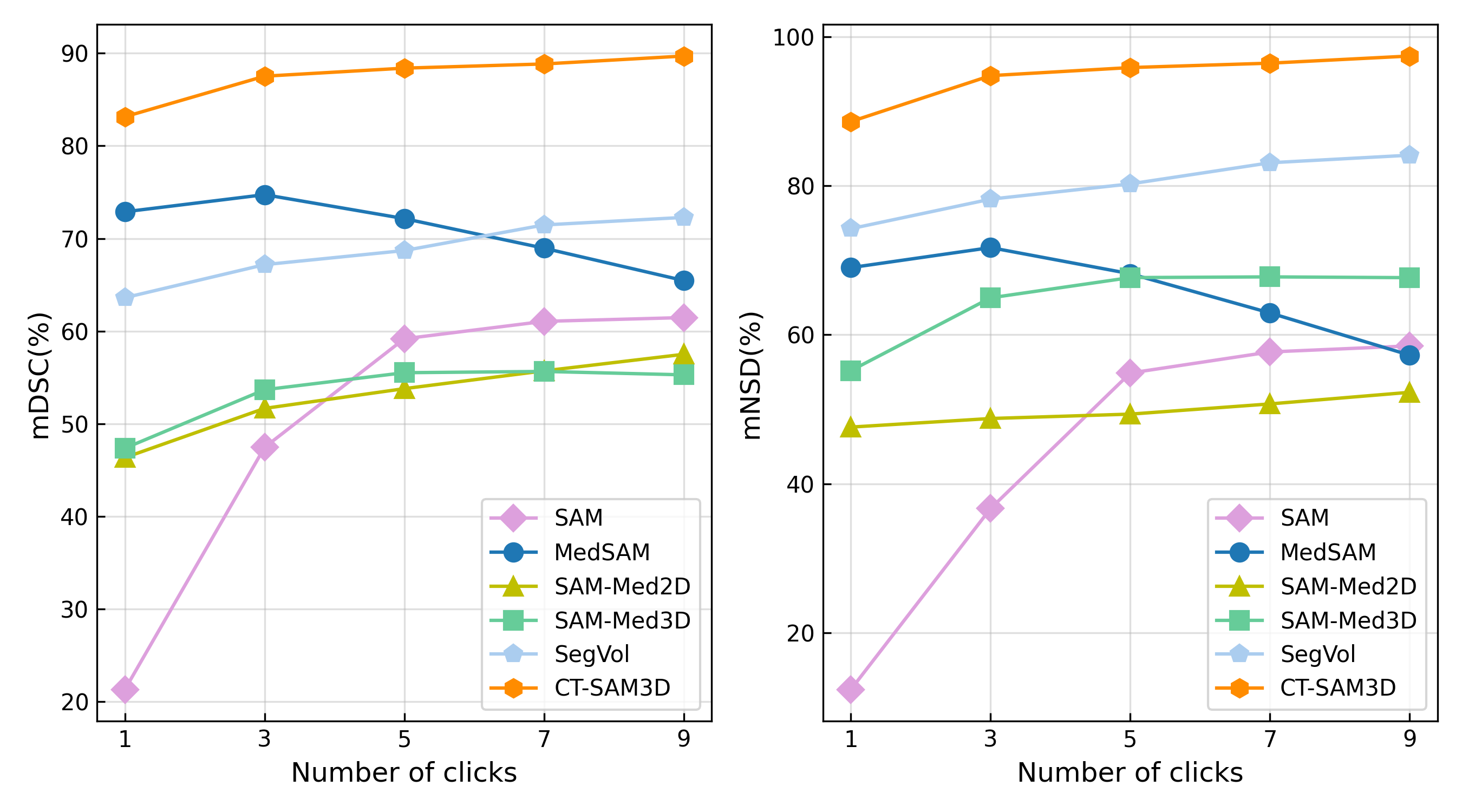}
\vspace{-2em}
\caption{\small Results under increasing number of clicks on FLARE22.}
\label{fig:flare}
\vspace{-1em}
\end{figure}

\begin{figure*}[t]
\centering
\includegraphics[width=1.0\linewidth]{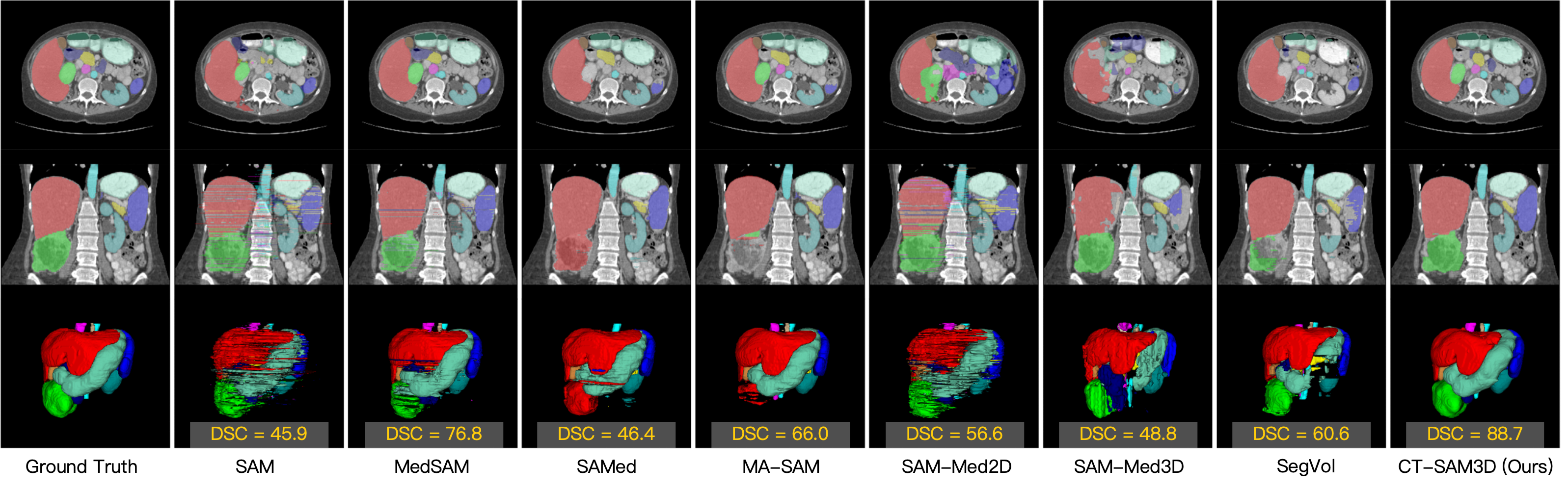}
\vspace{-2em}
\caption{\small Qualitative results of different methods on a subject who exhibits severe renal pathology (\textcolor{green}{green} region). The first row is an axial slice, the second row is a coronal slice, and the last row shows the 3D volume rendering. DSC (\%) scores are mentioned for each method.}
\label{fig:ab_vis}
\vspace{-1em}
\end{figure*}

\subsection{Main Results}
\noindent \textbf{Internal testing on TotalSeg++} is summarized in~\cref{fig:totalseg}. We found that only SAM-Med3D and SegVol can achieve meaningful results on over 100 anatomies during testing, partly because TotalSeg is a subset of their training sets. Therefore, only these two methods are taken as baselines in this experiment.  
We group the annotated anatomies into 9 groups (head, excretory organs, cardiovascular vessels, main chest organs, main abdomen organs, vertebrae, ribs, muscles, and other bones).
Compared to SAM-Med3D and SegVol, our CT-SAM3D yields substantially better performance across all groups. 
%Among them, the ribs group exhibits the largest performance improvements with median DSC scores increased by $\sim$50\% to $\sim$70\%. 
These results suggest that simply replacing 2D operations in SAM with its 3D counterparts does not work well, although the TotalSeg dataset is already included in their training set. In contrast, CT-SAM3D uses only $1139$ CT scans for training and achieves evidently superior performance than SAM-Med3D and SegVol, by proposing more suitable 3D SAM network architecture with effective prompt encoding.
To investigate the impact of the newly added annotations, we build a degraded CT-SAM3D* that is trained only on TotalSeg. As observed, the median DSC of most groups in CT-SAM3D surpasses that of CT-SAM3D*, and several groups (ribs, muscles, and other bones) achieve statistically significant improvements ($p<0.05$), indicating the newly added annotations have a positive impact on our model. Moreover, CT-SAM3D* still performs significantly better than almost all groups in SegVol and SAM-Med3D, indicating the superiority of our model architecture. Detailed results for each anatomy, including the newly added annotations, are provided in the supplementary material.

%\subsection{Quantitative Results of  Testing}
\noindent \textbf{External testing on FLARE22} is summarized in~\cref{tbl:quant_flare} and ~\cref{fig:flare}. ~\cref{tbl:quant_flare} presents the detailed organ-wise segmentation results when the prompt number $N=5$, while~\cref{fig:flare} illustrates the organ-averaged segmentation accuracy with respect to the prompt number $N$, as $N$ varies from 1 to 9. As shown in~\cref{tbl:quant_flare} and~\cref{fig:flare}, the proposed CT-SAM3D generalizes well to the external testing and performs significantly better as compared to other SAM-derived models with a large margin of $> \sim$10\% DSC and $> \sim$15\% NSD. Several interesting observations can be made. 1) The very recent 3D SAM models (SAM-Med3D and SegVol) do not generalize well, although they are trained on $21K$ and $6K$ medical images respectively. 2) Although SAMed and MA-SAM disable the prompt encoding module, they achieve the 2nd and 3rd best accuracy (if not considering the missing targets). This demonstrates the effectiveness of adaptation methods to some degree. However, these adaptation methods target on segmenting only a small fixed number of organs, limiting their general interactive segmentation ability. 3) We see from~\cref{fig:flare} that with more point prompts, CT-SAM3D's segmentation accuracy continues to increase, illustrating its interactive segmentation capacity. In contrast, with even 9 prompts per input sample, other methods still obtain undesirable segmentation accuracy, ranging from 55\% to 75\% DSC or from 50\% to 85\% NSD. 4) Not all methods improve consistently when the number of clicks increases, e.g., MedSAM starts to degrade when $N > 3$, indicating its deficiency in understanding/encoding user's prompts. A qualitative example is shown in~\cref{fig:ab_vis}.

\noindent \textbf{External testing on BTCV} is illustrated in~\cref{tbl:btcv}. It can be seen that with only 1 click prompt, CT-SAM3D has a mDSC of 78.4\% and mNSD of 88.4\%, substantially higher than all other interactive SAM-derived models. When increasing the number of clicks to 3, CT-SAM3D improves with $3.8\%$ mDSC (from 78.4\% to 82.2\%) and $5.5\%$ mNSD (from 88.4\% to 93.9\%). The obtained performance (3 clicks per organ) is also comparable to the fully supervised segmentation model trained on BTCV. These results demonstrate CT-SAM3D's generalization ability to unseen datasets, indicating the effectiveness for potential real clinical usage.

 \begin{table}[t]
 {\begin{tabular}{lcccc}
    \toprule[1pt]
        \multicolumn{1}{c}{\multirow{2}{*}{Methods}} & \multicolumn{2}{c}{1 click} & \multicolumn{2}{c}{3 clicks} \\ \cmidrule{2-3} \cmidrule{4-5}
        \multicolumn{1}{c}{} & \multicolumn{1}{c}{mDSC$\uparrow$} & \multicolumn{1}{c}  {mNSD$\uparrow$} & \multicolumn{1}{c}{mDSC$\uparrow$} & \multicolumn{1}{c}  {mNSD$\uparrow$} \\
    \midrule
    SAM & 14.2 & 6.8 & 39.3 & 31.3 \\
    MedSAM & 64.2 & 61.8 & 67.9 & 67.8\\
    SAM-Med2D & 42.3 & 51.3 & 48.1 & 48.3 \\
    \midrule
    SAM-Med3D & 41.9 & 51.5 & 47.5 & 61.2\\
    SegVol & 59.5 & 71.3 & 63.4 & 76.2 \\
    \rowcolor{gray!20}
     \textbf{CT-SAM3D} & \textBF{78.4} & \textBF{88.4} & \textBF{82.2}  & \textBF{93.9}\\ 
    \bottomrule[1pt]
\end{tabular}
}
\vspace{-0.5em}
{\caption{\upshape \small Mean DSC (\%) and NSD (\%) results on BTCV dataset.}\label{tbl:btcv}}
\end{table}

\begin{table}[t]
\centering
\setlength{\tabcolsep}{1.0mm}{
\begin{tabular}{lcccc}
    \toprule[1pt]
        \multicolumn{1}{c}{\multirow{2}{*}{Methods}} & \multicolumn{2}{c}{Pancreas Tumor} & \multicolumn{2}{c}{Colon Cancer} \\ \cmidrule{2-3} \cmidrule{4-5}
        \multicolumn{1}{c}{} & \multicolumn{1}{c}{mDSC$\uparrow$} & \multicolumn{1}{c}  {mNSD$\uparrow$} & \multicolumn{1}{c}{mDSC$\uparrow$} & \multicolumn{1}{c}  {mNSD$\uparrow$} \\
    \midrule
    nnU-Net & 41.65 & 62.54 & 43.91 & 52.52 \\
    nnFormer & 36.53 & 53.97 & 24.28 & 32.19 \\
    Swin UNETR & 40.57 & 60.05 & 35.21 & 42.94\\
    UNETR++ & 37.25 & 53.59 & 25.36 & 30.68\\
    3D UX-Net & 34.83 & 52.56 & 28.50 & 32.73\\
    \midrule
    SAM & 30.55 & 32.91 & 39.14 & 42.70 \\
    3DSAM-adapter & 57.47 & \cellcolor{gray!20}\textBF{79.62} & 49.99 & \cellcolor{gray!20}\textBF{65.67}\\
     \textbf{CT-SAM3D} & \cellcolor{gray!20}\textBF{59.60} & 77.93 & \cellcolor{gray!20}\textBF{50.68} & {64.14}\\ 
    \bottomrule[1pt]
\end{tabular}}
\vspace{-0.5em}
\caption{\upshape \small Zero-shot testing of CT-SAM3D on tumor segmentation.}
\label{tbl:tumor_seg}
\vspace{-1em}
\end{table}

\noindent \textbf{Zero-shot tumor segmentation.} 
Beyond anatomical structure segmentation tasks, we have applied CT-SAM3D to more challenging endeavors, i.e., tumor segmentation, to investigate its zero-shot capabilities. A previous SAM adaptation method, 3DSAM-adapter~\cite{gong20233dsam}, has also reported results on the aforementioned two tumor datasets. 3DSAM-adapter are fine-tuned on each dataset, followed by testing on a randomly selected 20\% of the data. Our significant distinction is that we do not conduct any fine-tuning operation, resulting in a truly zero-shot testing. For a fair comparison, we report results on the same test splits used in 3DSAM-adapter. The results are summarized in~\cref{tbl:tumor_seg}. Some state-of-the-art automatic segmentation methods are also included~\cite{isensee2021nnu, zhou2023nnformer, tang2022self, shaker2022unetr++, lee20223d}. As observed, a promptable, human-in-the-loop approach can greatly elevate the upper-bound of challenging tumor segmentation results. CT-SAM3D outperforms nnU-Net by 17.95\% mDSC on pancreas tumor segmentation, primarily benefiting from input prompts that directly indicate the location of the tumors. When compared to the specifically fine-tuned 3DSAM-adapter, our CT-SAM3D achieves comparable performance on both pancreas tumor segmentation and colon cancer segmentation tasks under 10 clicks, maintaining a slight edge on mDSC yet exhibiting a minor shortfall in mNSD, even under a more challenging zero-shot setting. A qualitative comparison is shown in our supplementary material, where we showcase results of SAM, 3DSAM-adapter, and our proposed CT-SAM3D under gradually increased click prompts. This comparison further demonstrates the excellent zero-shot capability of CT-SAM3D.

\subsection{Ablation Study Results}
{\bf The effectiveness of PSAP.} To find an effective prompt encoding method, we have carefully compared the performance of random Fourier features (RFF)~\cite{rahimi2007random, tancik2020fourier} and our proposed progressively and spatially aligned prompt (PSAP) on the tasks of FLARE22 and BTCV. It is observed that, under the same image encoder (ResT), PSAP achieves improvements of 20.7\% and 21.1\% in mDSC and mNSD, respectively, compared to RFF on BTCV, as shown in~\cref{tbl:quant_psap}. Similar observations are obtained on FLARE22. Interestingly, when we replace the image encoder with the UNet~\cite{cciccek20163d} structure, which is more common in medical image analysis, we can still obtain very competitive results. This implies that an effective prompt encoding mechanism is essentially crucial in 3D interactive segmentation. It also reveals that PSAP can serve as a plug-and-play module to be incorporated into different hierarchical backbone network structures to construct a stronger interactive segmentation model.

\noindent {\bf The effectiveness of CPP.} To investigate the effectiveness of cross-patch prompt (CPP) on segmenting large organs, we conduct a 1-click experiment on liver and aorta segmentation tasks (\cref{tbl:quant_cpp}). As expected, the results are quite disappointing without CPP due to large organ dimensions. Yet, a single click with enabled CPP mechanism can boost mDSC performance of liver by 45.2\% and 42.4\% on FLARE22 and BTCV, respectively. Even on a tubular-shaped aorta structure, CPP can bring $\sim$10\% gains both in mDSC and mNSD scores. It should also be noted that there is still room for further improvement as the number of clicks increases.

\begin{table}[t]
%\addtolength{\leftskip} {-1cm} 
\setlength{\tabcolsep}{0.5mm}{
\begin{tabular}{lcc|cccc}
    \toprule[1pt]
    \multicolumn{1}{c}{\multirow{2}{*}{I.Enc.}} & \multicolumn{1}{c}{\multirow{2}{*}{w/ RFF}} & \multicolumn{1}{c}{\multirow{2}{*}{w/ PSAP}} & 
    \multicolumn{2}{c}{FLARE22} & \multicolumn{2}{c}{BTCV} \\ 
    \cmidrule{4-5} \cmidrule{6-7}
        \multicolumn{1}{c}{} & \multicolumn{1}{c}{} & \multicolumn{1}{c}{} & \multicolumn{1}{c}{mDSC$\uparrow$} & \multicolumn{1}{c}  {mNSD$\uparrow$} & \multicolumn{1}{c}{mDSC$\uparrow$} & \multicolumn{1}{c}  {mNSD$\uparrow$} \\
        
    \midrule
    ResT & \checkmark & & 73.1 & 81.4 & 61.5 & 72.8\\
    ResT & & \checkmark & \cellcolor{gray!20}\textBF{87.5} & \cellcolor{gray!20}\textBF{94.8}& \cellcolor{gray!20}\textBF{82.2} & \cellcolor{gray!20}\textBF{93.9}\\
    UNet & & \checkmark & 86.4 & 93.8 & 79.9 & 92.0\\
    \bottomrule[1pt]
\end{tabular}}
\vspace{-0.5em}
\caption{\upshape \small Effectiveness of PSAP. ``I.Enc.'' stands for image encoder.}
\label{tbl:quant_psap}
\end{table}

\begin{table}[t]
\centering
\setlength{\tabcolsep}{1.0mm}{
\begin{tabular}{lc|cccc}
    \toprule[1pt]
     \multicolumn{1}{c}{\multirow{2}{*}{Organ}} & \multicolumn{1}{c}{\multirow{2}{*}{w/ CPP}} & \multicolumn{2}{c}{FLARE22} & \multicolumn{2}{c}{BTCV} \\ 
    \cmidrule{3-4} \cmidrule{5-6}
        \multicolumn{1}{c}{}  & \multicolumn{1}{c}{} & \multicolumn{1}{c}{mDSC$\uparrow$} & \multicolumn{1}{c}  {mNSD$\uparrow$} & \multicolumn{1}{c}{mDSC$\uparrow$} & \multicolumn{1}{c}  {mNSD$\uparrow$} \\
    \midrule
    Liver & & 41.9 & 31.5 & 38.1 & 29.7 \\
    Liver & \checkmark & \cellcolor{gray!20}\textBF{87.1} & \cellcolor{gray!20}\textBF{77.1} & \cellcolor{gray!20}\textBF{80.5} & \cellcolor{gray!20}\textBF{70.3}\\
    \midrule
    Aorta & & 51.9 & 55.7 & 46.8 & 51.1\\
    Aorta & \checkmark & \cellcolor{gray!20}\textBF{61.6} & \cellcolor{gray!20}\textBF{64.1} & \cellcolor{gray!20}\textBF{57.7} & \cellcolor{gray!20}\textBF{61.9}\\
    
    \bottomrule[1pt]
\end{tabular}}
\vspace{-0.5em}
\caption{\upshape \small Effectiveness of CPP on large organs under 1 click.}
\label{tbl:quant_cpp}
\vspace{-1em}
\end{table}

% \begin{figure}[ht]
% \centering
% \includegraphics[width=0.6\linewidth]{ablation.png}
% \caption{\small The qualitative comparisons about the ablation studies of PSAP and CPP, where the \textcolor{green}{green} dot denotes the user click.}
% \label{fig:ablation}
% \vspace{-1em}
% \end{figure}

%The effectiveness of the proposed progressively and spatially aligned prompt (PSAP) and cross-patch prompt (CPP) modules are illustrated in~\cref{fig:ablation}. Without the PSAP, the point prompt may not be effectively encoded. Therefore, there is no segmentation mask predicted. Enhanced with PSAP, CT-SAM3D successfully produces a reasonable segmentation mask for the skeletal muscle. 
%The second row shows a large organ (liver) segmentation example. If there is no CPP training, one-click input can segment a local ROI with maximum range of the training patch size ($64\times64\times64$). Hence, for large organs, like the liver in this example, multiple click inputs are required to segment a complete 3D liver. In contrast, CT-SAM3D can produce a reasonable whole liver segmentation using only one click. Due to the space limit, please refer to our supplementary material for the quantitative ablation results.

%% file: sec/5_conclusion.tex
\section{Conclusion}
\label{sec:conclusion}
In this work, we present a comprehensive, efficient and 3D promptable model on whole-body CT scans. Instead of adapting SAM, we directly develop a pure 3D promptable model utilizing a more comprehensively labeled CT dataset (i.e., TotalSeg++). To train CT-SAM3D effectively using 3D local image patches, we propose two key technical developments to encode the click prompt in local 3D space and conduct the cross-patch prompt scheme to reduce clicks when segmenting large organs. CT-SAM3D significantly outperforms all previous SAM-derived models by a large margin and demonstrates strong zero-shot capability.

%% file: sec/X_suppl.tex
\clearpage
\maketitlesupplementary
%\hypersetup{linkcolor=eccvblue}
% \tableofcontents
%\hypersetup{linkcolor=red}

\section{Data Curation Process}
\label{sec:dcp}
Using an in-house developed muscle/fat segmentation model with manual examination and curation, we enhance the TotalSeg~\cite{wasserthal2023totalsegmentator} dataset by introducing annotations of three important yet under-explored anatomies, i.e., skeletal muscle, visceral fat, and subcutaneous fat. This results in a more complete whole-body CT dataset where overall $\sim$83\% of voxels within the body region are semantically labeled. To accurately determine this annotated ratio, we utilize a body region prediction model to exclude areas of background, defined as regions with a Hounsfield Unit (HU) of -1024 and regions corresponding to the CT table in the imaging data. Subsequently, we cumulate the count of annotated pixels within the body across the entire dataset to ascertain the overall proportion of intracorporeal annotations. More specifically on the curation process of the newly added three anatomies, we use the in-house muscle/fat segmentation model to predict the skeletal muscle, visceral fat, and subcutaneous fat regions for all scans in TotalSeg dataset. 
Since a portion of the muscles, such as the iliopsoas, have already been annotated in TotalSeg, we retain the original labels for these specific regions.
Then, two radiologists with over five years of experience manually refine the boundaries between the newly added annotations and existing structures. We refer to the enhanced dataset as TotalSeg++, for CT-SAM3D developing and evaluation.

\section{Hyper-parameters and Network Details}
We summarize the hyper-parameters and network details in~\cref{tbl:hyper_params}. One thing that needs a detailed explanation is how we obtain samples $U$ and $V$ with overlapping regions during training. Concretely, we first sample a specific anatomical label from the label set of the input scan. Then, we randomly select an anchor point from all the foreground points of this anatomical structure. Centering on this point, we allow the centers of $U$ and $V$ to randomly deviate from the anchor point by up to half the patch size in each direction. As a result, the distance between the centers of $U$ and $V$ on any axis falls within the range of $[0, 64]$, when patch size is set as $(64 \times 64 \times 64)$. We also elaborate on the network architecture configuration and the computational analysis in~\cref{tbl:hyper_params}. It can be observed that the light-weight CPP prediction network $\mathcal{P}$ adds only a minor computational overhead (1.3G FLOPs) compared to the primary interactive segmentation network $\mathcal{S}$ (124.4G FLOPs).

\begin{table}[t]
\centering
%\addtolength{\leftskip} {-1cm} 
\setlength{\tabcolsep}{0.6mm}{
\begin{tabular}{l|c}
    \toprule[1pt]
    Configuration & Value\\
    \midrule
    AdamW $\beta$ & $(0.9, 0.999)$ \\
    Weight decay & $1e{-2}$ \\
    Initial learning rate & $1e{-4}$ \\
    Warmup epochs & 100 \\
    Training epochs & 1000 \\
    Batch size on single GPU & 4 \\
    Number of GPUs & 8 \\
    Number of samples per volume & 8 \\
    Number of iterations per batch & 5 \\
    \midrule
    Patch size & $(64 \times 64 \times 64)$ \\
    Center distance range between $U$ and $V$ & $[0, 64]$\\
    \#Params of segmentation network $\mathcal{S}$ &  72.5M  \\
    FLOPs of $\mathcal{S}$ & 124.4G\\
    Embedding dimensions in 4 stages of $\mathcal{S}$ & [96, 192, 384, 768] \\
    Number of ResT blocks in 4 stages of $\mathcal{S}$ & [1, 2, 6, 2] \\
    \#Params of CPP prediction network $\mathcal{P}$ &  4.8M\\  
    FLOPs of $\mathcal{P}$ & 1.3G\\
    \midrule
    NSD Tolerance &  5 \\
    
    \bottomrule[1pt]
\end{tabular}}
\caption{\upshape Hyper-parameters and computational parameters.}
\label{tbl:hyper_params}
\end{table}

\section{Quasi-Real-Time 3D Interactive Tool}
\begin{figure*}[t]
\centering
\includegraphics[width=0.8\linewidth]{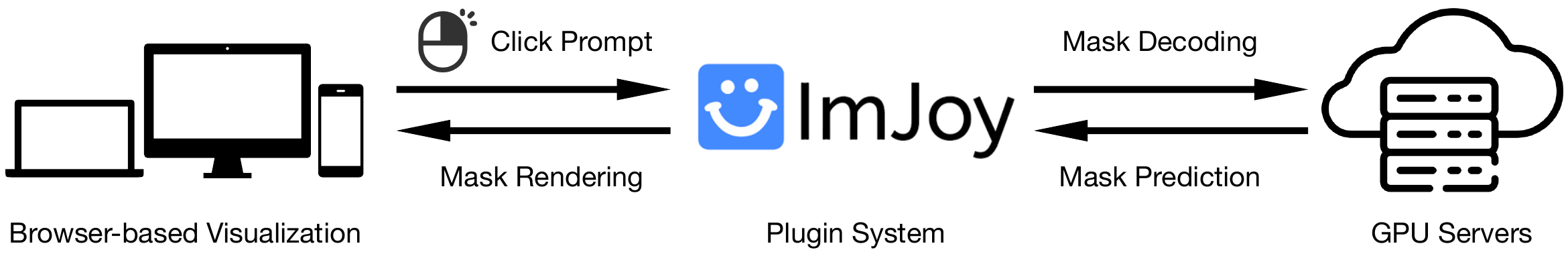}
\caption{The diagram of our interactive segmentation tool.}
\label{fig:interactive_tool}
\end{figure*}

Classic medical imaging interaction tools, such as ITK-SNAP~\cite{yushkevich2016itk} and 3D Slicer~\cite{pieper20043d}, have been instrumental in propelling the development of the field. Nonetheless, for 3D interactive segmentation tasks, these platforms often necessitate meticulous slice-by-slice manipulation to obtain fine-grained segmentation outcomes, indicating substantial potential for efficiency enhancements. On the other hand, interactive models like SAM~\cite{Kirillov_2023_ICCV} are becoming increasingly intelligent nowadays, and how to efficiently run such models on these platforms remains an open problem. Therefore, we seek to build our 3D interactive segmentation tool based on a flexible and open-source browser-based platform, namely ImJoy~\cite{ouyang2019imjoy}, to facilitate the developing and validation of our proposed CT-SAM3D. ImJoy can access various computational resources by performing computations either in the
browser itself or using ``plugins'' that can run either locally or remotely. Building upon this, we embed the 3D image and mask visualization directly within the browser, which also receives user click prompts, while launching the model inference computation to GPU servers for enhanced efficiency. An illustrative diagram is shown in~\cref{fig:interactive_tool}. With this efficient 3D interactive segmentation tool, users can experience response times of less than one second for each interaction cycle. This rapid feedback is incredibly beneficial for practical clinical use, allowing for seamless and efficient workflows.

\section{Additional Experimental Results}
\noindent {\bf Comparison with automatic segmentation methods.}

\noindent Several cutting-edge automatic segmentation methods, including nnU-Net~\cite{isensee2021nnu}, UNETR~\cite{hatamizadeh2022unetr}, Swin UNETR~\cite{tang2022self}, nnFormer~\cite{zhou2023nnformer} and TotalSeg Model~\cite{wasserthal2023totalsegmentator} are compared in this section to provide a full picture about the positioning of CT-SAM3D in current medical image segmentation field. This comparison is performed on the FLARE22 dataset, and results are summarized in~\cref{tbl:auto_seg}. Compared with the fully automatic methods, interactive CT-SAM3D has higher numerical values on quantitative metrics. For instance, compared with nnU-Net, CT-SAM3D has a 4.8\% improvement in mDSC. It is important to clarify that we are not intending to claim superiority of interactive methods over fully automatic ones. Indeed, an effective interactive method is expected to utilize prompts and error regions to enhance segmentation refinement. We wish to draw attention to a different perspective and the practical value that emerges from this comparison: a promptable, human-in-the-loop collaborative approach can further boost segmentation performance and step towards real-world clinical applications.

\begin{table}[t]
\centering
\setlength{\tabcolsep}{1.0mm}{
\begin{tabular}{lcc}
    \toprule[1pt]
        Methods & {mDSC$\uparrow$} & {mNSD$\uparrow$} \\
    \midrule
    nnU-Net~\cite{isensee2021nnu} & 83.6 & 88.5   \\
    UNETR~\cite{hatamizadeh2022unetr} & 85.8 & 90.0  \\
    Swin UNETR~\cite{tang2022self} & 86.8 & 91.2 \\
    nnFormer~\cite{zhou2023nnformer} & 86.9 & 91.3\\
    TotalSeg Model~\cite{wasserthal2023totalsegmentator} & 86.5 & 94.1 \\ 
    \midrule
     \textbf{CT-SAM3D} & \cellcolor{gray!20}\textBF{88.4} & \cellcolor{gray!20}\textBF{95.8}\\ 
    \bottomrule[1pt]
\end{tabular}}
\caption{\upshape Quantitative comparison with leading automatic segmentation methods on FLARE22 dataset.}
\label{tbl:auto_seg}
\end{table}

\noindent {\bf Qualitative comparison on tumor segmentation.}

\begin{figure*}[!t]
\centering
\includegraphics[width=0.93\linewidth]{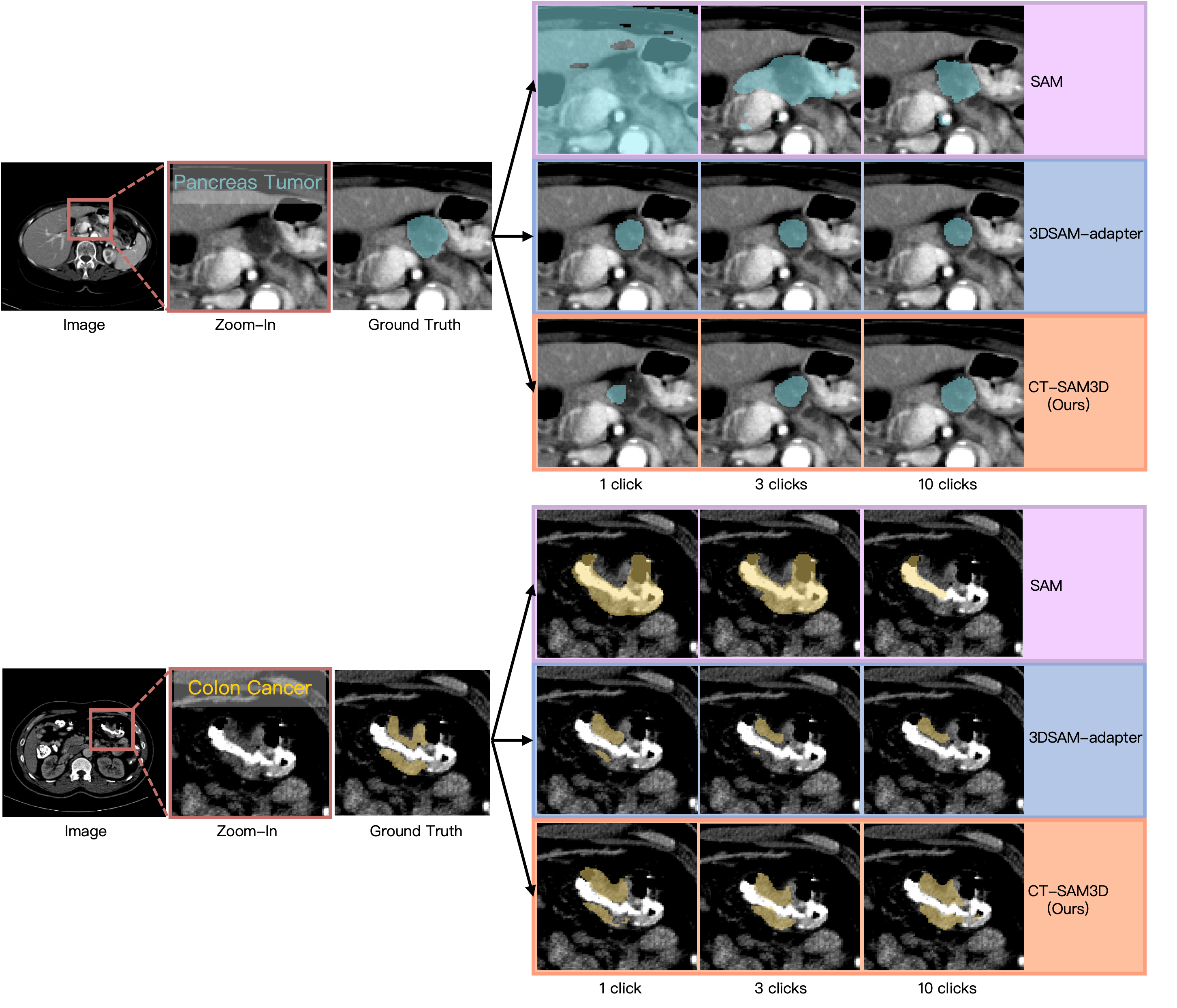}
\caption{Qualitative visualizations of pancreas tumor and colon cancer segmentation. We showcase results of SAM, 3DSAM-adapter, and our proposed CT-SAM3D under gradually increased click prompts. CT-SAM3D will substantially approach the ground truth masks as the number of clicks increases. This comparison prominently demonstrates the excellent zero-shot capability of CT-SAM3D. Best viewed in color.}
\label{fig:tumor_visual}
\end{figure*}

\noindent We have presented quantitative zero-shot tumor segmentation results in the main text. We supplement in this material the qualitative tumor segmentation results, as shown in~\cref{fig:tumor_visual}. 
Two cases from MSD-Pancreas and MSD-Colon with pancreatic cancer and colon cancer are elaborately visualized.
We showcase results of SAM~\cite{Kirillov_2023_ICCV}, 3DSAM-adapter~\cite{gong20233dsam}, and our proposed CT-SAM3D under gradually increased click prompts. It can be seen that the results of SAM have a more obvious improvement with the increase of the number of clicks, but even the results of 10 clicks are still quite different from the ground truth masks. As for 3DSAM-adapter, since it uses MSD-Pancreas and MSD-Colon data for fine-tuning, its results will hardly improve with the change of the number of clicks, which seems to indicate that fine-tuning for a specific dataset will cause the model to lose its interactive or zero-shot ability. In contrast, CT-SAM3D will substantially approach the ground truth masks as the number of clicks increases.
This qualitative comparison further demonstrates the excellent zero-shot capability of CT-SAM3D.

\begin{table}[t]
\centering
%\addtolength{\leftskip} {-1cm} 
\setlength{\tabcolsep}{0.8mm}{
\begin{tabular}{l|l}
    \toprule[1pt]
        Anatomy Group & Anatomy Labels\\
    \midrule
    Head & 50, 93 \\
    Excretory Organs & 2, 3, 55, 57, 104 \\
    Cardiovascular Vessels & 7, 8, 9, 49, 51--54 \\
    Main Chest Organs & 13--17, 42--48 \\
    Main Abdomen Organs & 1, 4, 5, 6, 10, 11, 12, 56\\
    Vertebrae  & 18--41\\
    Ribs & 58--81\\
    Other Bones & 82--92\\
    Muscles & 94--103\\
    Skeletal Muscles (Curated) & 105 \\
    Visceral Fat (Curated)& 106 \\
    Subcutaneous Fat (Curated) & 107 \\
    \bottomrule[1pt]
\end{tabular}}
\caption{\upshape Grouped label lists of TotalSeg++.}
\label{tbl:label_group}
\end{table}

\noindent {\bf Qualitative results of TotalSeg++.}

\begin{figure*}[!t]
\centering
\includegraphics[width=0.93\linewidth]{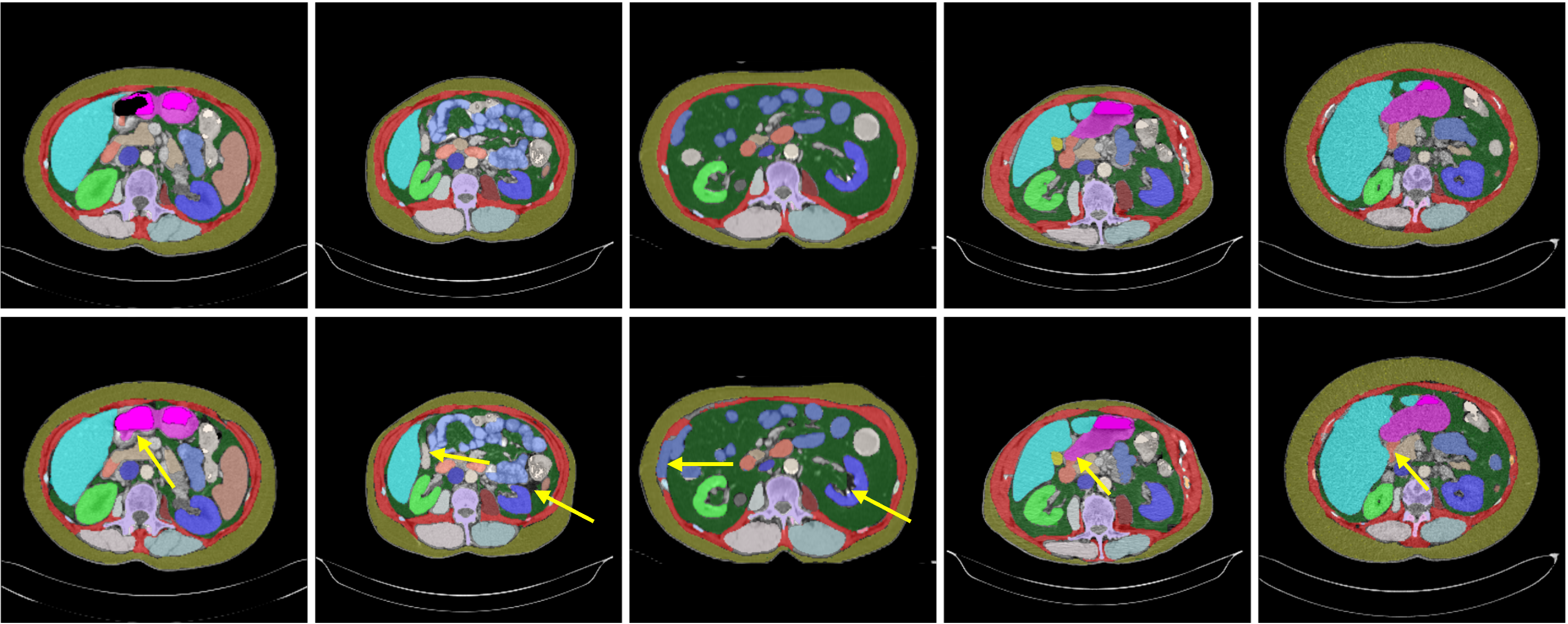}
\caption{Qualitative results of CT-SAM3D on TotalSeg++ testing set. The first row shows ground truths and the second row shows our merged prediction results. Overall, our method demonstrates promising consistency with ground truths, except for a few discrepancies indicated by golden arrows. Best viewed in color.}
\label{fig:totalseg_visual}
\end{figure*}

\noindent In~\cref{fig:totalseg_visual}, we provide more qualitative results of our CT-SAM3D on anatomical structure segmentation task. From this figure, it is evident that our method can not only segment major organs effectively but also handle extensive regions of skeletal muscle, visceral fat, and subcutaneous fat with proficiency. 
Overall, our method exhibits strong agreement with the ground truths, with only minor discrepancies arising in a few subtle areas.
This indicates that it can provide fundamental support for downstream tasks such as body composition analysis and sarcopenia assessment, playing a practical role in clinical settings.

\noindent {\bf Detailed quantitative results of TotalSeg++.}

\begin{figure*}[t]
\centering
\includegraphics[width=1.2\linewidth, angle=90]{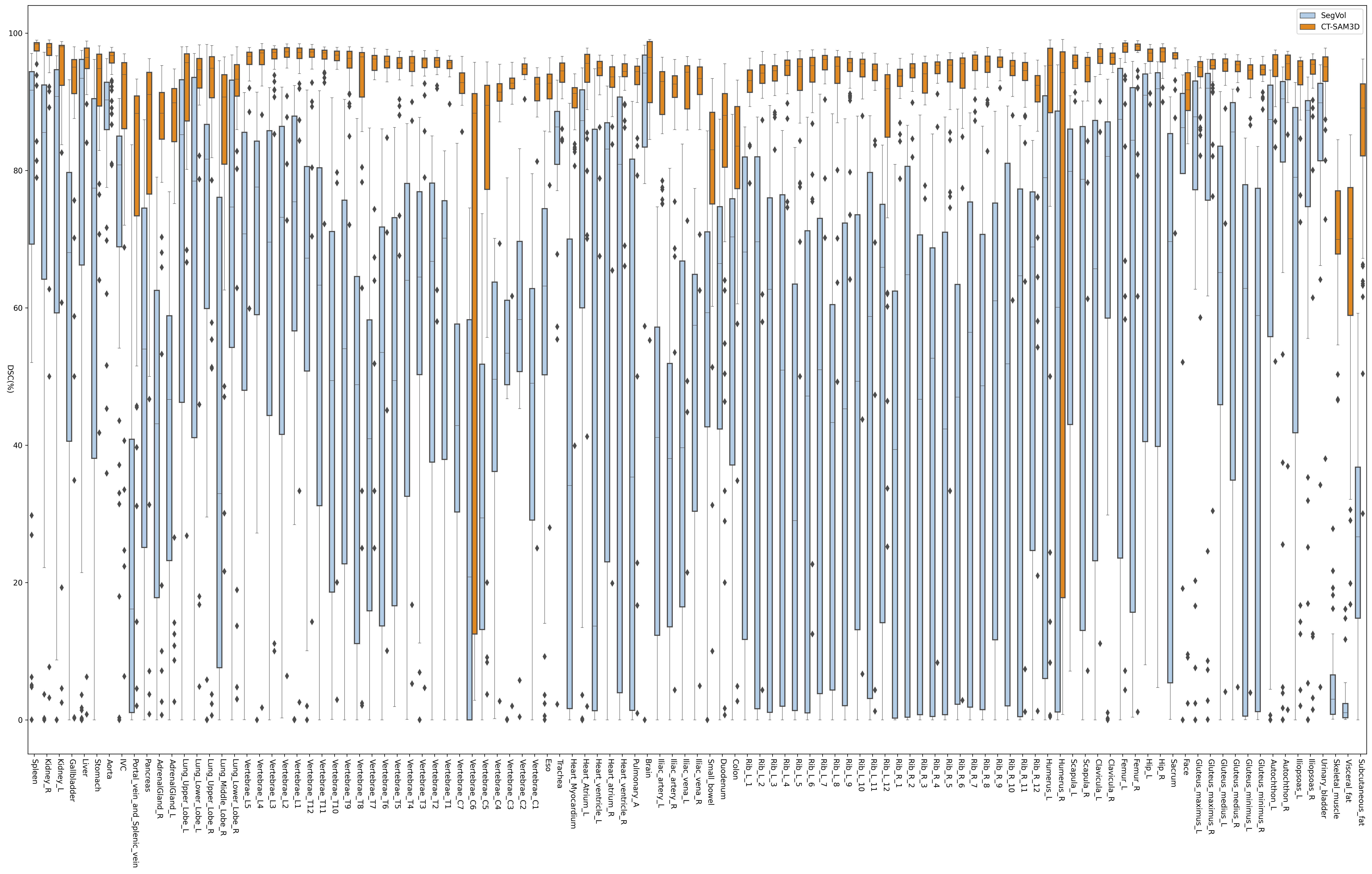}
\caption{DSC results of all 107 anatomic structures annotated in TotalSeg++. Best viewed when rotated 90 degrees.}
\label{fig:totalseg_full_dsc}
\vspace{-1em}
\end{figure*}

\begin{figure*}[t]
\centering
\includegraphics[width=1.2\linewidth, angle=90]{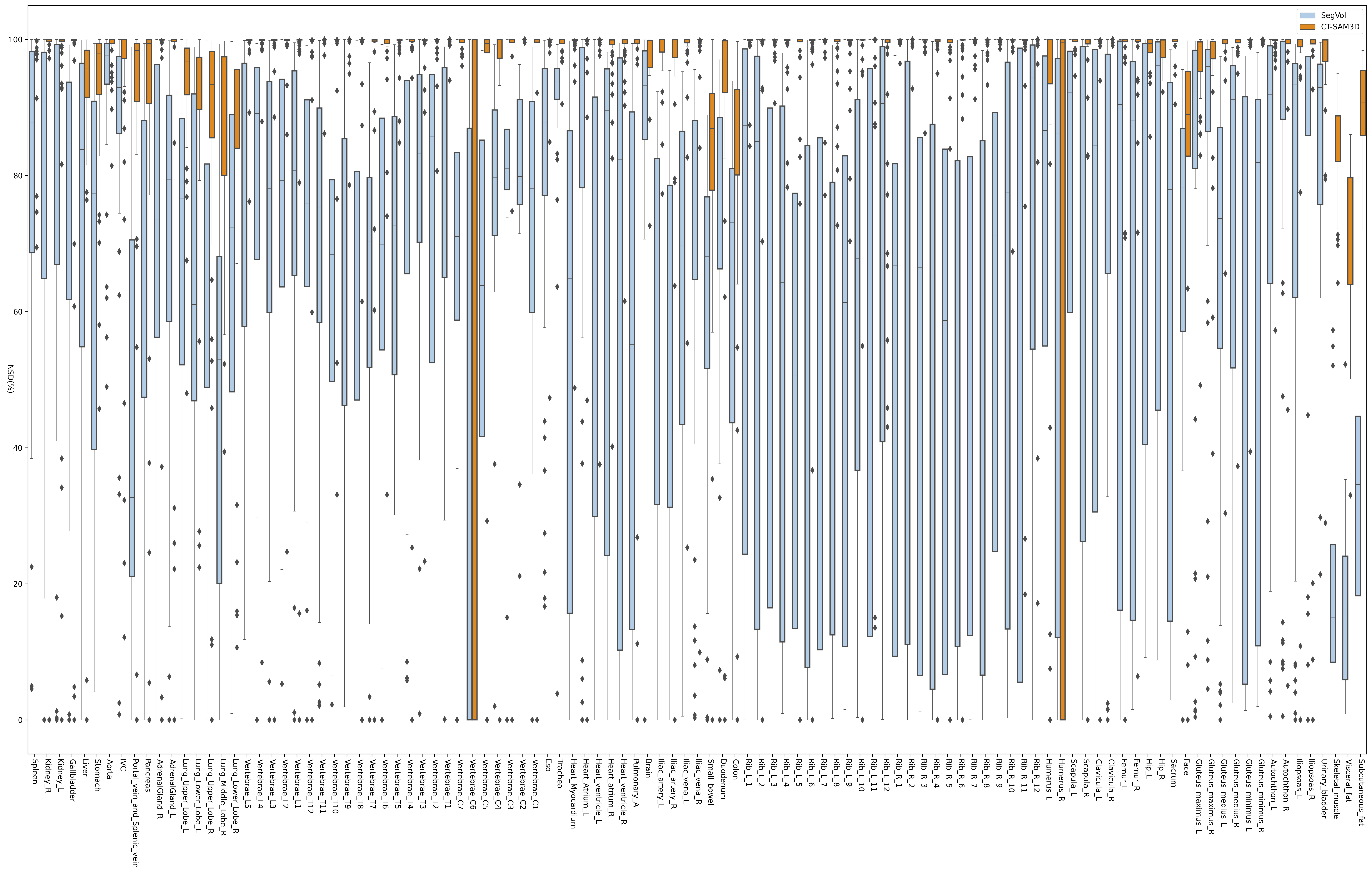}
\caption{NSD results of all 107 anatomic structures annotated in TotalSeg++. Best viewed when rotated 90 degrees.}
\label{fig:totalseg_full_nsd}
\vspace{-1em}
\end{figure*}

\noindent Due to space limitations, it is infeasible to report every single organ performance in the main text as there is a total of 107 anatomical structures in TotalSeg++. Hence, we group the original 104 organs of TotalSeg into 9 groups (head, excretory organs, cardiovascular vessels, main chest organs, main abdomen organs, vertebrae, ribs, muscles, and other bones) and use our added three important yet under-explored anatomies (curated skeletal muscles, visceral fat, and subcutaneous fat) as another 3 separate categories, and report the group-wise DSC in the main text. The label list of each group is summarized in~\cref{tbl:label_group}.
In this supplementary material, we also provide a detailed complement to the results for all 107 anatomic structures in the test set of TotalSeg++, as shown in~\cref{fig:totalseg_full_dsc} and ~\cref{fig:totalseg_full_nsd}. As observed, our CT-SAM3D significantly outperforms SegVol~\cite{du2023segvol} (the overall second-best 3D SAM method) across all structures in both DSC and NSD metrics.

\noindent {\bf Efficiency comparison.}

\noindent We have measured computing resource-related metrics and consulted with our collaborating radiologists about the approximate time consumption for manual correction of a single organ and annotation using our CT-SAM3D (baseline methods have not been integrated into our interactive system). The results are compiled into~\cref{tbl:efficiency}. It clearly demonstrates a significant boost in efficiency. Of note, latency and GPU memory measurements obtained on different hardware may exhibit some variations.

\begin{table*}[b]
\centering
\setlength{\tabcolsep}{1.0mm}{
\begin{tabular}{lllllll}
    \toprule[1pt]
        Methods & {Input Size} & {Params (M)} & GFLOPs & Latency & Memory (MB) & Time/Organ\\
    \midrule
    SAM & 1024$\times$1024 & 93.74 & 370.63 & 0.38 $s/slice$ & 3128 & - \\
    MedSAM & 1024$\times$1024 & 93.74 & 370.63 & 0.33 $s/slice$ & 3116 & - \\
    SAMed & 224$\times$224 & 92.19 & 103.08 & 0.14 $s/slice$ & 1376 & - \\
    MA-SAM & 512$\times$512 & 96.81 & 885.18 & 0.42 $s/slice$ & 7230 & - \\
    SAM-Med2D & 256$\times$256 & 271.24 & 43.54 & 0.36 $s/slice$ & 1582 & - \\
    SAM-Med3D & 128$\times$128$\times$128 & 100.51 & 89.85 & 0.31 $s/patch$ & 2798 & -\\
    SegVol & 128$\times$128$\times$128 & 117.72 & 181.76 & 0.18 $s/patch$ & 4472 & - \\
    \textbf{CT-SAM3D} & 64$\times$64$\times$64 & 77.30 & 125.70 & 0.14 $s/patch$ & 1240 & $<$ 10 $s$\\
    Manual (SliceThickness $>$ 3 mm) & - & - & - & - & - & $\sim$ 0.5 $h$ \\
    Manual (SliceThickness $\le$ 3 mm) & - & - & - & - & - & $\sim$ 1-2 $h$ \\
    \bottomrule[1pt]
\end{tabular}}
\caption{\upshape A comparative analysis of the efficiency of different methods, including manual editing with the approximate time consumption listed in the table for reference.}
\label{tbl:efficiency}
\end{table*}

\section{Limitations}
Although the proposed CT-SAM3D has achieved appealing results on a large variety of anatomic structures and tumors, there still exist several limitations. 

Firstly, our model may have unsatisfactory results for extremely incomplete structures presented in the scan. E.g., the 6th cervical vertebra (Vertebrae\_C6) and the right humerus (Humerus\_R) show unstable performance in~\cref{fig:totalseg_full_dsc}. Upon further examination, it has been found that a substantial percentage of the scans containing these two categories have corresponding annotation regions fewer than 100 voxels, posing a great challenge for our model.

Secondly, since our annotations do not contain different levels of granularity, prompting anatomic structures with finer granularity may require additional interactions. E.g., in the case of the liver segmentation, we only have annotations for the entire liver to train our model. If a user wishes to prompt a specific Couinaud segment of the liver, it would necessitate more interactions.

Lastly, our current system is not yet capable of automatically extracting semantic information of anatomic structures, which will be addressed in our future work.